\pdfoutput=1
\documentclass[10pt,twocolumn,letterpaper]{article}

\usepackage[pagenumbers]{wacv}
\usepackage{amsmath,amssymb,mathtools}
\usepackage{algorithm}
\usepackage{algpseudocode}
\usepackage{booktabs}
\usepackage{graphicx}
\usepackage{microtype}
\usepackage{multirow}
\usepackage{placeins}
\usepackage{xcolor}

\newcommand{\R}{\mathbb{R}}

\usepackage{cuted}

\definecolor{wacvblue}{rgb}{0.21,0.49,0.74}
\usepackage[breaklinks,colorlinks,allcolors=wacvblue]{hyperref}
\hypersetup{
    pdftitle={LE-GDA: Layer Ensemble Gaussian Discriminant Analysis for Training-Free Few-Shot CLIP Adaptation},
    pdfauthor={Pavodi Maniamfu et al.}
}

\def\confName{WACV}
\def\confYear{2027}

\title{LE-GDA: Layer Ensemble Gaussian Discriminant Analysis for Training-Free
Few-Shot CLIP Adaptation}

\author{\parbox{0.95\textwidth}{\centering\normalsize
Pavodi Maniamfu$^{1}$ \quad Vogel Kiketa$^{2,3}$\\
Zezina Kuzulu Kuetu$^{2}$ \quad Jakin Mukendi Kabongo$^{4,2}$\\
David Mayoya Kutangila$^{2,3}$ \quad Pescie Honore Kinkani Mampuya$^{3}$\\
L'Or Wolo Sieka$^{5}$\\[4pt]
{\small
$^{1}$ Temple University Japan Campus; $^{2}$ Universit\'e de Kinshasa;\\
$^{3}$ Universit\'e Protestante au Congo; $^{4}$ Handong Global University;\\
$^{5}$ Kwilu Briques
}
\\[3pt]{\small\textit{Paper accepted at WACV 2027}}
\\[2pt]{\small\textit{Corresponding author:} \texttt{ndoyi.pavodi.maniamfu [at] tuj.temple.edu}}
}}

\begin{document}
\maketitle

\begin{abstract}
Training-free few-shot adaptation of vision--language models typically
constructs a classifier from CLIP's final projected image representation,
leaving intermediate representations unused. Combining them is not straightforward. Class scores from different representations differ in spread by orders of magnitude for the same query. Direct averaging allocates influence by numerical scale rather than predictive information. We propose Layer-Ensemble Gaussian Discriminant Analysis (LE-GDA), a method that fits regularized shared-covariance discriminants independently to intermediate and final representations extracted in one frozen encoder pass.
LE-GDA standardizes each discriminant's class scores per query, selects a small
representation multiset on labeled validation data, and combines it with the
zero-shot classifier without gradient-based optimization. Across 11 datasets,
four CLIP backbones, five shot counts, and three independently sampled $K$-shot
training sets, LE-GDA achieves the highest mean accuracy of all evaluated
methods on every backbone, raising the 16-shot mean from 78.01\% to 79.54\%
over a matched GDA-CLIP (final-feature) baseline and improving that baseline
for every backbone--shot combination. At one shot, LE-GDA retains a 1.20-point
mean gain over the baseline even though covariance estimation is unavailable and per-query standardization is neutral. A control with richer per-class structure inside a single representation does not reproduce the gain, indicating that the benefit comes from combining representations. These results show that intermediate CLIP representations carry information beyond the final embedding and can improve training-free few-shot adaptation. Code and evaluation resources are available at \url{https://github.com/pavodi-NM/le-gda}.
\end{abstract}
\section{Introduction}
\label{sec:intro}

Vision-language models such as Contrastive Language-Image Pre-training (CLIP)~\cite{clip} learn image representations
that transfer effectively to downstream recognition tasks. In zero-shot
classification, predictions are obtained through the cosine similarity between the encoded image features and text embeddings generated from class-specific prompt templates. For example, a template like ``a photo of \{class\}'' allows the language encoder to construct the necessary classifier weights. This enables the model to categorize images across arbitrary domains without downstream fine-tuning. To further improve performance, a small few-shot training set can introduce
task-specific information through prompt learning, feature adapters, residual
classifiers, or non-parametric retrieval as adopted by recent efficient fine-tuning methods
\cite{coop,clipadapter,taskres,tipadapter}.  Although these approaches can be
parameter-efficient, methods that optimize task-specific parameters still
require a training procedure for every downstream dataset.  Training-free
adaptation~\cite{tipx_2023_ICCV, proker_2025_CVPR, ape, protoadapter, gdaclip, gpcache} instead constructs the classifier directly from frozen CLIP
features.

In this work, we study training-free adaptation in the validation-assisted regime. This is the convention adopted by cache-based CLIP adapters~\cite{tipadapter, ape, gdaclip, gpcache}, where classifiers are fit on the few support examples per class. A held-out labeled split selects the fusion and regularization hyperparameters these methods require. Within this regime, a gain is only interpretable if every method draws on the same validation budget~\cite{clap}.

Gaussian discriminant analysis provides a strong training-free classifier for
CLIP~\cite{gdaclip}. It estimates class statistics from the few-shot training features and combines the resulting linear discriminant with the zero-shot classifier. Like cache- and adapter-based methods more generally~\cite{tipadapter,clipadapter,gdaclip}, it operates on the final projected image embedding.  This representation is the natural default
because it is aligned with the text encoder and is often the most accurate
individual representation.  It is nevertheless only the last output of a
sequence of transformations performed by the visual encoder.

Combining intermediate representations requires more than naive classifier averaging.  Each representation defines a different statistical feature space, with its own dimensionality, covariance structure, and per-query dispersion of scaled discriminant-scores. The last is the spread of a representation's class scores within a single query and varies by more than three orders of magnitude across the intermediate representations of a single backbone (e.g. ResNets). Under direct score averaging, influence is therefore allocated by numerical
scale rather than predictive information. Normalizing scores before combination
is long-established practice when fusing classifiers whose outputs live on
different scales~\cite{kittler1998combining,jain2005score}. Training-free
CLIP adapters have not required it, since they operate in a single feature
space. A selection procedure operating on
such scores inherits the same bias, preferring representations for their scale
rather than their contribution to classification.

Building on~\cite{gdaclip}, which we refer to as GDA-CLIP, we propose Layer-Ensemble Gaussian Discriminant Analysis (LE-GDA) to combine
these representations. The method extracts intermediate and final
representations from a single frozen CLIP forward pass and fits the same
shared-covariance Gaussian discriminant to each representation. It standardizes the class scores of each representation within each query,
then greedily selects~\cite{caruana2004ensemble} a small subset of representations using validation
accuracy and combines their scores with the
zero-shot classifier. All classifiers are obtained from
few-shot training statistics, without back-propagation.

Our contributions are as follows:
\begin{itemize}
\item We show that CLIP's intermediate representations carry information a training-free ensemble can extract without gradient-based optimization. The same shared-covariance discriminant is fitted independently in each frozen feature space.

\item We identify per-query score standardization as the component that makes discriminants from different feature spaces commensurate under a single fusion coefficient. Validation selection therefore tracks predictive quality rather than scale.

\item We evaluate the method across 11 datasets, four CLIP backbones, and five
shot counts against a matched GDA-CLIP (final-feature) baseline. Controls
isolate the source of the gain. At one shot, where neither covariance estimation nor
standardized fusion applies, LE-GDA retains a
1.20-point mean gain.
\end{itemize}
\section{Related work}
\label{sec:related}

\noindent\textbf{Few-shot adaptation of vision--language models.}
CLIP~\cite{clip} supports zero-shot recognition by aligning images
and text in a shared embedding space. It nevertheless underperforms in
specialized and domain-shifted settings~\cite{clap,coop}. This motivates adaptation from a limited
labeled subset. CoOp~\cite{coop} introduced learnable, continuous prompt tokens into the language branch to optimize the textual context. CLIP-Adapter~\cite{clipadapter} learns residual feature transformations
through MLPs, TaskRes~\cite{taskres} optimizes a residual classifier, and ClusterAdapter~\cite{clusteradapter} learns multiple prototypes per class by gradient descent. These methods obtain task-specific parameters through gradient-based training.

Training-free approaches instead construct predictors directly from frozen features. CLAP~\cite{clap} distinguishes validation-dependent from validation-free
adaptation and shows that access to labeled validation data is an important
part of the evaluation protocol. LP++~\cite{lpplusplus} similarly reduces
task-specific validation by deriving its optimization step sizes from the
training objective.
Tip-Adapter~\cite{tipadapter} stores few-shot training examples in a key-value
cache,
while Proto-Adapter~\cite{protoadapter} uses a single average prototype per class instead.  GPCache
\cite{gpcache} applies Gaussian-process weight and confidence calibration to cache retrieval.  These methods differ in how few-shot training information is represented, but generally operate on the final CLIP image embedding. Separately, APE~\cite{ape} selects informative channels within the final feature representation.  We study whether frozen intermediate
representations can contribute to training-free adaptation.

\noindent\textbf{Gaussian discriminant analysis and covariance estimation.}
Gaussian discriminant analysis models class-conditional features with
Gaussian distributions and yields a linear classifier when the covariance is
shared across classes~\cite{bishop2006prml}.  GDA-CLIP~\cite{gdaclip} applies
this classifier to the final CLIP image embedding.   Estimating a precision matrix from high-dimensional
few-shot features is ill-conditioned, motivating covariance shrinkage and
regularization~\cite{ledoitwolf2004,kubokawa2008}.  GDA-CLIP uses an
empirical-Bayes ridge estimator.  We retain the shared-covariance classifier,
use a validation-selected trace-scaled ridge coefficient, and fit the
classifier independently in several feature spaces.

\noindent\textbf{Intermediate representations.}
Intermediate neural representations have long been used to improve
dense prediction and optimization. Examples include hypercolumn
features~\cite{hypercolumns} and deep supervision~\cite{deeplysup}.
Recent work exploits CLIP's intermediate layers for few-shot
classification. It does so through trained side adapters and
learned multi-scale fusion~\cite{multilayertuning,metafeature}.
These approaches introduce gradient-optimized modules on top of the
retained features. We instead fit the same closed-form discriminant
independently in each frozen feature space. To our knowledge, training-free
few-shot adaptation of CLIP has so far operated exclusively on the
final projected embedding.
\section{Layer-ensemble Gaussian discriminant analysis}
\label{sec:method}

\subsection{Problem formulation and feature spaces}

For a task with $C$ classes and $K$ labeled training examples per class, let
$\mathcal S=\bigcup_{c=1}^{C}\mathcal S_c$ denote the $K$-shot training set and
$N=CK$.  The frozen CLIP text encoder produces normalized text embeddings
$W=[w_1,\ldots,w_C]\in\R^{D\times C}$, where $D$ is the dimension of CLIP's
shared image--text embedding space.  Given an image $x$, which we refer to as a
query at inference time, a single pass through the frozen CLIP image
encoder yields the normalized final embedding $f_{\mathrm{last}}(x)$ and
zero-shot logits
\begin{equation}
a(x)=\tau f_{\mathrm{last}}(x)^\top W,
\qquad \tau=100.
\label{eq:zs}
\end{equation}

We retain intermediate representations from that same pass. For vision transformer (ViT)~\cite{vit} backbones, the candidate set contains the
normalized class token (CLS) after transformer blocks 4, 6, 8, 10,
and 11. It also includes the feature after the final layer normalization and before projection, together with the final projected embedding.  For ResNet~\cite{resnet} backbones, we apply global
average pooling after each of the four residual stages and include the final
projected embedding.  We index the candidate feature spaces by $\mathcal L$ and denote
the normalized feature in space $\ell$ by
$f_\ell(x)\in\R^{D_\ell}$.

\subsection{Gaussian discriminant in each feature space}

LE-GDA builds on the classifier of GDA-CLIP~\cite{gdaclip}. That method
estimates class means and a shared covariance from the few-shot training set.
It converts these statistics into a linear discriminant on the final CLIP image
embedding, then combines the resulting scores with the zero-shot classifier. We
retain both the shared-covariance discriminant and the zero-shot fusion. Our
change is to fit that discriminant independently to each candidate
representation, before feature-space selection and score fusion.

For each space $\ell$, we assume class-conditional Gaussian features with a
covariance matrix shared across classes,
\begin{equation}
p\!\left(f_\ell(x)\mid y=c\right)
=\mathcal N\!\left(\mu_{c,\ell},\Sigma_\ell\right),
\quad p(y=c)=C^{-1}.
\label{eq:gaussian}
\end{equation}
Here, $y\in\{1,\ldots,C\}$ is the class label, $c$ indexes a class,
$\mathcal N(\mu,\Sigma)$ denotes a multivariate Gaussian distribution,
$\mu_{c,\ell}\in\R^{D_\ell}$ is the mean of class $c$ in feature space
$\ell$, and $\Sigma_\ell\in\R^{D_\ell\times D_\ell}$ is the covariance
matrix shared by all classes in that space.  The prior $p(y=c)=C^{-1}$ assigns
equal probability to every class.  Under this model, Bayes' rule gives a
linear discriminant after removing terms that are constant across
classes~\cite{bishop2006prml}.  We estimate the normalized class mean as
\begin{equation}
\mu_{c,\ell}=
\frac{\sum_{x_i\in\mathcal S_c}f_\ell(x_i)}
{\left\lVert\sum_{x_i\in\mathcal S_c}f_\ell(x_i)\right\rVert_2},
\label{eq:mean}
\end{equation}
where $\mathcal S_c$ contains the $K$ labeled training images from class $c$,
$x_i$ denotes one such image.  Thus, $\mu_{c,\ell}$ is the unit-normalized
average feature for class $c$ in space $\ell$.
\begin{align}
\widehat{\Sigma}_\ell &=\frac{1}{N}\sum_{c=1}^{C}\sum_{x_i\in\mathcal S_c}
\bigl(f_\ell(x_i)-\mu_{c,\ell}\bigr)
\bigl(f_\ell(x_i)-\mu_{c,\ell}\bigr)^\top,
\label{eq:cov}\\
\Sigma_\ell &=\widehat{\Sigma}_\ell+\lambda_\ell
\frac{\operatorname{tr}(\widehat{\Sigma}_\ell)}{D_\ell}I.
\end{align}
Here, $\widehat{\Sigma}_\ell\in\R^{D_\ell\times D_\ell}$ is the pooled within-class
covariance estimate centered on the normalized class mean,
following~\cite{gdaclip}; $\Sigma_\ell$ is the regularized covariance estimate
used by the classifier.  The factor
$\operatorname{tr}(\widehat{\Sigma}_\ell)/D_\ell$ is the average marginal variance, and
$\lambda_\ell\geq0$ controls the strength of the ridge regularization.  The
class score is
\begin{equation}
g_{\ell,c}(x)=f_\ell(x)^\top\Sigma_\ell^{-1}\mu_{c,\ell}
-\frac{1}{2}\mu_{c,\ell}^\top\Sigma_\ell^{-1}\mu_{c,\ell}.
\label{eq:gda}
\end{equation}
Here, $g_{\ell,c}(x)$ is the discriminant score assigned to class $c$ for
image $x$ in space $\ell$, and $\Sigma_\ell^{-1}$ is the corresponding
precision matrix.  We collect the $C$ class scores into
$g_\ell(x)=[g_{\ell,1}(x),\ldots,g_{\ell,C}(x)]^\top\in\R^C$.  The equal
class prior contributes the same constant to every score and is therefore
omitted.  The ridge coefficient $\lambda_\ell$ is selected independently for
each space on validation data.  Specifically, for $K>1$ we choose
$\lambda_\ell$ from the ridge grid $\Lambda$ to maximize the standalone
top-1 accuracy
$\operatorname{Acc}_{\mathcal V}(x\mapsto\arg\max_c
g_{\ell,c}^{(\lambda)}(x))$ on
$\mathcal V$, before zero-shot fusion and representation selection.  When
$K=1$, every within-class residual is zero, so no covariance can be estimated.
We set $\Sigma_\ell=I$, reducing
\Cref{eq:gda} to a nearest-class-mean classifier.  After fitting,
\Cref{eq:gda} is stored as a linear head, so the covariance matrix is
needed only during fitting and is not retained at inference.

\subsection{Per-query score standardization}

Different feature spaces produce discriminant vectors with different
query-dependent offsets and scales.  Before fusion, we standardize the scores
of each space across the $C$ classes for every query:
\begin{equation}
q_{\ell,c}(x)=
\frac{g_{\ell,c}(x)-\bar g_\ell(x)}
{\sqrt{\frac{1}{C}\sum_{j=1}^{C}
\left(g_{\ell,j}(x)-\bar g_\ell(x)\right)^2+\delta_v}}.
\label{eq:zscore}
\end{equation}
Here, $q_{\ell,c}(x)$ is the standardized score for class $c$,
$\bar g_\ell(x)=C^{-1}\sum_{j=1}^{C}g_{\ell,j}(x)$ is the mean score across
classes for the query image, and $\delta_v>0$ regularizes the variance; we use
$\delta_v=10^{-12}$.  We
write $q_\ell(x)=[q_{\ell,1}(x),\ldots,q_{\ell,C}(x)]^\top\in\R^C$ for the
standardized score vector.  The transformation preserves the class ordering
within each space and is exactly invariant to additive shifts
$g_\ell(x)\mapsto g_\ell(x)+\beta\mathbf 1$.  When $\delta_v=0$, it is also
exactly invariant to positive scalar rescaling of $g_\ell(x)$.  The fixed
$\delta_v$ implies a standard-deviation floor $\sqrt{\delta_v}=10^{-6}$ in
score units, so scale invariance holds above this floor.

\subsection{Feature-space selection and fusion}
\label{sec:selection_fusion}

Let $\mathcal V$ denote the labeled validation set and $\mathcal A$ the set of
candidate values for the fusion coefficient.  We select at most $T=4$ feature
spaces by greedy forward selection with replacement on $\mathcal V$.  At each
step, every candidate is appended to the current multiset and scored with the
best value in $\mathcal A$; the candidate with the highest validation accuracy
is retained.  Among the four prefixes we keep the most accurate on $\mathcal
V$, giving the selected multiset $\mathcal L^*$.  Repetition provides a
discrete weighting within a mixture, without introducing feature-specific
continuous coefficients.  It also reduces overfitting of the selection
procedure itself~\cite{caruana2004ensemble}.  This is ensemble selection with
the candidate set formed by varying the representation rather than the
learning algorithm.

The final logits are
\begin{equation}
z(x)=a(x)+\frac{\alpha}{|\mathcal L^*|}
\sum_{\ell\in\mathcal L^*}q_\ell(x).
\label{eq:fusion}
\end{equation}
Here, $z(x)\in\R^C$ is the final logit vector, $a(x)\in\R^C$ is the zero-shot
logit vector from \cref{eq:zs}, $|\mathcal L^*|$ is the number of elements in
the selected multiset including repetitions, and $\alpha\in\mathcal A$ is the
validation-selected coefficient controlling the contribution of the visual
discriminants.

For the non-standardized baseline, we use
$\widehat g_\ell(x)=g_\ell(x)/r_\ell$, where
$r_\ell=\max\{|\operatorname{median}_{u\in\mathcal V}\max_c
g_{\ell,c}(u)|,10^{-12}\}$; the supplementary material gives the full definition in \cref{eq:supp_scale}.

By contrast, \cref{eq:fusion} with $q_\ell$
is invariant to positive rescaling of $g_\ell$ above the standard-deviation
floor in \cref{eq:zscore}, since greedy selection and the $\alpha$ search operate on
standardized scores alone.  The method estimates class means and shared
covariances from the $K$-shot training set.  It selects only $\lambda_\ell$,
$\mathcal L^*$, and $\alpha$ on validation data.  No parameter is updated by
gradient descent.

Relative to the baseline GDA-CLIP~\cite{gdaclip}, LE-GDA therefore introduces
three changes.  These are multiple representations, per-representation ridge
selection, and standardized fusion of a selected ensemble.
\Cref{sec:experiments} describes the matched baseline that isolates them.  To
use one covariance convention throughout, both procedures in
\cref{alg:comparison} define
$\widehat{\Sigma}$ with the $1/N$ normalization in \cref{eq:cov}.  At
$\lambda=D/(N-1)$, the ridge precision satisfies
\begin{align}
&D\bigl((N-1)\widehat{\Sigma}+\operatorname{tr}(\widehat{\Sigma})I\bigr)^{-1}
\notag\\[-2pt]
&\quad=\frac{D}{N-1}
\left(\widehat{\Sigma}+\frac{\operatorname{tr}(\widehat{\Sigma})}{N-1}I\right)^{-1},
\label{eq:gda_correspondence}
\end{align}
whose left-hand side is the published GDA-CLIP precision.  The prefactor is
absorbed by $\alpha$, so \cref{alg:comparison} writes the right-hand side
without it.

\begin{algorithm*}[t]
\caption{GDA-CLIP~\cite{gdaclip} and LE-GDA in the common ridge parameterization
of \cref{eq:gda_correspondence}.  GDA-CLIP uses the final embedding; LE-GDA
standardizes and validation-selects an ensemble from one encoder pass.}
\label{alg:comparison}
\begin{minipage}[t]{0.475\textwidth}
\textbf{(a) GDA-CLIP}
\vspace{2pt}\hrule\vspace{3pt}
\footnotesize
\begin{NoHyper}
\begin{algorithmic}[1]
\Require $K$-shot training set $\mathcal S$, validation set $\mathcal V$,
query $x$,
text classifier $W$, fusion grid $\mathcal A$
\State Extract final normalized features $f_{\mathrm{last}}$ from the frozen
image encoder
\For{$c=1,\ldots,C$}
  \State $\mu_c \gets \operatorname{Normalize}\!\left(
  \sum_{x_i\in\mathcal S_c}f_{\mathrm{last}}(x_i)\right)$
\EndFor
\State Estimate $\widehat{\Sigma}$ with the $1/N$ normalization in \cref{eq:cov}
\If{$K=1$}
  \State $P\gets I$
\Else
  \State $P\gets\left(\widehat{\Sigma}
  +\operatorname{tr}(\widehat{\Sigma})I/(N-1)\right)^{-1}$
\EndIf
\For{$c=1,\ldots,C$}
  \State $H_c\gets P\mu_c$;
  $b_c\gets-\tfrac12\mu_c^\top P\mu_c$
\EndFor
\State $g(u)\gets f_{\mathrm{last}}(u)^\top H+b$
\State $\alpha\gets\arg\max_{\gamma\in\mathcal A}
\operatorname{Acc}_{\mathcal V}\!\left(a+\gamma g\right)$
\State \Return $\arg\max_c\left[a(x)+\alpha g(x)\right]_c$
\end{algorithmic}
\end{NoHyper}
\end{minipage}
\hfill
\begin{minipage}[t]{0.505\textwidth}
\textbf{(b) LE-GDA}
\vspace{2pt}\hrule\vspace{3pt}
\footnotesize
\begin{NoHyper}
\begin{algorithmic}[1]
\Require $K$-shot training set $\mathcal S$, validation set $\mathcal V$,
query $x$,
text classifier $W$, candidate spaces $\mathcal L$, ridge grid $\Lambda$,
fusion grid $\mathcal A$
\State Extract and normalize $\{f_\ell\}_{\ell\in\mathcal L}$ from one
frozen image-encoder pass
\ForAll{$\ell\in\mathcal L$}
  \For{$c=1,\ldots,C$}
    \State $\mu_{c,\ell}\gets
    \operatorname{Normalize}\!\left(\sum_{x_i\in\mathcal S_c}
    f_\ell(x_i)\right)$
  \EndFor
  \If{$K=1$}
    \State $\Sigma_\ell\gets I$
  \Else
    \State Estimate $\widehat{\Sigma}_\ell$
    \State Select $\lambda_\ell\in\Lambda$ by standalone
    $\operatorname{Acc}_{\mathcal V}$
    \State $\Sigma_\ell\gets \widehat{\Sigma}_\ell+\lambda_\ell
    \operatorname{tr}(\widehat{\Sigma}_\ell)I/D_\ell$
  \EndIf
  \For{$c=1,\ldots,C$}
    \State $H_{c,\ell}\gets\Sigma_\ell^{-1}\mu_{c,\ell}$;
    $b_{c,\ell}\gets-\tfrac12\mu_{c,\ell}^\top H_{c,\ell}$
  \EndFor
  \State $q_\ell(u)\gets
  \operatorname{Standardize}_c\!\left(f_\ell(u)^\top H_\ell+b_\ell\right)$
\EndFor
\State $(\mathcal L^*,\alpha)\gets
\operatorname{GreedySelect}_{\mathcal V}
\left(\{q_\ell\},\mathcal A,T=4\right)$ with replacement
\State $z(x)\gets a(x)+\dfrac{\alpha}{|\mathcal L^*|}
\sum_{\ell\in\mathcal L^*}q_\ell(x)$
\State \Return $\arg\max_c z_c(x)$
\end{algorithmic}
\end{NoHyper}
\end{minipage}
\end{algorithm*}
\section{Experimental setup}
\label{sec:experiments}

\paragraph{Datasets.}
We use the 11 datasets commonly adopted for few-shot CLIP classification:
Caltech101~\cite{feifei2004caltech101}, DTD~\cite{cimpoi2014dtd},
EuroSAT~\cite{helber2019eurosat}, FGVC-Aircraft~\cite{maji2013aircraft},
Flowers102~\cite{nilsback2008flowers}, Food101~\cite{bossard2014food101},
ImageNet~\cite{deng2009imagenet}, OxfordPets~\cite{parkhi2012pets},
StanfordCars~\cite{krause2013cars}, SUN397~\cite{xiao2010sun}, and
UCF101~\cite{soomro2012ucf101}.  Ten datasets use the standard test splits from the CoOp~\cite{coop}
evaluation protocol.

\paragraph{ImageNet protocol.}
The CoOp protocol provides no held-out test split for ImageNet. The official
validation set is conventionally used for evaluation instead. LE-GDA selects
its ridge coefficients, feature-space multiset, and fusion coefficient on
validation data. Evaluating on the full validation set would therefore place
model-selection images inside the evaluation set.\footnote{In the reference
implementation of GDA-CLIP~\cite{gdaclip}, the ImageNet branch assigns the
validation features to the test features, so the fusion coefficient $\alpha$ is
selected on the same 50{,}000 images used for evaluation. The other ten
datasets use a disjoint validation split.} We instead split the official
validation images within each class into 10 for model selection and 40 for
evaluation. The split is class-balanced. Every method in
\Cref{tab:backbone_main} uses that partition, the same sampled $K$-shot
training sets, and the same cached features. Our ImageNet accuracies are
therefore computed on a class-balanced 40,000-image subset. They are not
directly comparable with published results over all 50,000 validation images.

\paragraph{Backbones and few-shot training sets.}
We evaluate CLIP with ResNets~\cite{resnet} RN50 and RN101 and vision transformer~\cite{vit} ViT-B/32 and ViT-B/16 as visual encoder backbones.  For each $K$, we sample
three $K$-shot training sets using three random seeds and average the resulting
runs.  The complete evaluation uses
$K\in\{1,2,4,8,16\}$ labeled images per class.  The main tables use $K=16$,
while the shot-count evaluation compares LE-GDA with its matched
GDA-CLIP (final-feature) baseline at every $K$.

\paragraph{Validation protocol.}
Our primary evaluation is validation-assisted. Labeled validation images are
available for hyperparameter and representation selection. Classifiers are fit
only on the $K$ training examples per class. This is the validation-dependent
regime in the terminology of CLAP~\cite{clap}. For the ten non-ImageNet
datasets we use the fixed validation partition from the CoOp
splits~\cite{coop}. During feature extraction this partition is class-balanced
and capped at $\max\{30,\lfloor 3000/C\rfloor\}$ images per class. Classes with
fewer examples retain all of them. ImageNet uses the 10 images per class
defined above. Validation images are disjoint from both the few-shot sampling
pool and the test set. The validation set is also unchanged across shot counts,
so $K=1$ draws on the same validation data as $K=16$. Every method that
requires model selection uses these images, with its own hyperparameter grid.

\paragraph{Baselines.}
We report zero-shot CLIP, Tip-Adapter~\cite{tipadapter}, GPCache
(W+C)~\cite{gpcache}, Tip-Adapter-F, Proto-Adapter-F~\cite{protoadapter}, and
our matched implementation of GDA-CLIP~\cite{gdaclip}, denoted GDA-CLIP
(final-feature). Our GPCache implementation includes weight and confidence
calibration. It omits similarity calibration, which trains a projection layer
on external unlabeled images. That module falls outside our protocol, which
uses no data beyond the $K$-shot training set and the validation split. Its
accuracy is therefore not comparable with the published GPCache figures. The
supplementary material reports an external consistency check against published
GPCache in \cref{tab:gpcache-published}, including an RN50 comparison with ImageNet excluded. We
exclude CLAP~\cite{clap} and LP++~\cite{lpplusplus}, whose validation-free
protocol would be changed by supplying labeled validation data. The trained
cache baselines use 20 optimization epochs. Our GDA-CLIP (final-feature)
baseline follows the original classifier~\cite{gdaclip}, selecting the
trace-scaled ridge and fusion coefficients on the same validation grids as
LE-GDA. By \cref{eq:gda_correspondence}, this ridge family contains the
published empirical-Bayes estimator at $\lambda=D_{\mathrm{last}}/(N-1)$ up to
grid resolution. Every baseline is reimplemented with the same $K$-shot
training sets, validation splits, cached features, text classifier, and
evaluation splits as LE-GDA. \Cref{tab:backbone_main} therefore reports
matched implementations rather than published numbers.

\paragraph{Hyperparameters.}
For every feature space, $\lambda_\ell$ is selected from 12 logarithmically
spaced values in $[10^{-8},10^2]$.  The coefficient $\alpha$ that combines
the visual discriminants with the zero-shot classifier is selected from 40
logarithmically spaced values in $[10^{-3},1.17\times10^5]$.  Greedy selection
uses replacement~\cite{caruana2004ensemble}, a maximum of four selections, and retains the best
validation prefix.  The final method uses the per-query standardization in
\cref{eq:zscore}, and no additional scale parameter is fitted for an
individual feature space.

\paragraph{Evaluation.}
Unless stated otherwise, all reported results average three independently
sampled $K$-shot training sets.  We also
report paired median differences over 11 datasets, four backbones, and three
random seeds, together with the number of dataset--backbone settings improved
after averaging the runs.  All reported hypothesis tests are two-sided; sign
tests use exact binomial probabilities after omitting ties.  Reported
$p$-values are unadjusted and descriptive.
\section{Results}
\label{sec:results}

\subsection{Comparison at 16 shots}

\Cref{tab:backbone_main} reports the mean accuracy over all 11 datasets for
each backbone. LE-GDA obtains the highest mean for every backbone. Relative to
the matched GDA-CLIP (final-feature) baseline, the per-backbone gains range
from 0.91 to 2.27 points (\cref{tab:shots}, $K=16$). Averaged over the four
backbones, LE-GDA reaches 79.54\%, compared with 78.01\% for GDA-CLIP
(final-feature) and 77.94\% for the strongest trained cache baseline. The
training-free GDA-CLIP head and the trained Tip-Adapter-F are therefore at
parity, differing by 0.07 points on the mean. LE-GDA exceeds both without
gradient-based optimization. After averaging the three runs, LE-GDA improves
the matched baseline in 36 of the 44 dataset--backbone settings. Across the
complete design of 11 datasets, four backbones, and three random seeds, the
median paired improvement is 0.829 points.
Figure~\ref{fig:fewshot-a} shows the corresponding curves for ViT-B/16 across
all shot counts.

Relative to the published GDA-CLIP averages, our matched GDA-CLIP
(final-feature) baseline is 0.37--0.65 points lower on every backbone,
consistent with our ImageNet partition (\cref{sec:experiments}).  LE-GDA
nevertheless exceeds those published averages on all four backbones, by
0.46--1.62 points.
\begin{table}[t]
\centering
\caption{Mean top-1 accuracy (\%) over 11 datasets at $K=16$. ``Grad.''
indicates gradient-based adaptation. ``Impl.'' indicates whether the row was
produced by the authors' released code (orig.) or by our implementation
(ours), in both cases under our protocol (Sec.~\ref{sec:experiments}). Bold denotes
the best result in each accuracy column.}
\label{tab:backbone_main}
\resizebox{\linewidth}{!}{%
\begin{tabular}{lccrrrrr}
\toprule
Method & Grad. & Impl. & RN50 & RN101 & ViT-B/32 & ViT-B/16 & Mean\\
\midrule
Zero-shot CLIP & -- & -- & 58.81 & 59.59 & 62.11 & 65.43 & 61.49\\
Tip-Adapter & -- & orig. & 70.20{\scriptsize$\pm$.13} & 72.06{\scriptsize$\pm$.12} & 72.88{\scriptsize$\pm$.16} & 76.88{\scriptsize$\pm$.02} & 73.01\\
Proto-Adapter & -- & ours & 72.48{\scriptsize$\pm$.19} & 73.64{\scriptsize$\pm$.14} & 74.52{\scriptsize$\pm$.03} & 78.34{\scriptsize$\pm$.06} & 74.74\\
GPCache (W+C) & -- & ours & 72.81{\scriptsize$\pm$.05} & 75.26{\scriptsize$\pm$.21} & 75.61{\scriptsize$\pm$.10} & 79.42{\scriptsize$\pm$.14} & 75.78\\
Tip-Adapter-F & \checkmark & orig. & 75.50{\scriptsize$\pm$.14} & 77.49{\scriptsize$\pm$.04} & 77.61{\scriptsize$\pm$.03} & 81.16{\scriptsize$\pm$.08} & 77.94\\
Proto-Adapter-F & \checkmark & ours & 75.39{\scriptsize$\pm$.10} & 76.67{\scriptsize$\pm$.09} & 76.76{\scriptsize$\pm$.02} & 80.39{\scriptsize$\pm$.06} & 77.30\\
GDA-CLIP (final-feature) & -- & ours & 75.56{\scriptsize$\pm$.12} & 77.41{\scriptsize$\pm$.18} & 77.69{\scriptsize$\pm$.09} & 81.40{\scriptsize$\pm$.06} & 78.01\\
\textbf{LE-GDA (ours)} & -- & ours & \textbf{77.47}{\scriptsize$\pm$.09} & \textbf{79.68}{\scriptsize$\pm$.25} & \textbf{78.69}{\scriptsize$\pm$.02} & \textbf{82.31}{\scriptsize$\pm$.04} & \textbf{79.54}\\
\bottomrule
\end{tabular}}
\end{table}

The improvement is broad but varies by dataset.  \Cref{tab:per_dataset}
averages each dataset over the four backbones.  LE-GDA improves nine of the 11
dataset averages, with the largest gains on EuroSAT and FGVC-Aircraft and the
two negative differences smaller than 0.2 points.  In a dataset-dropping
sensitivity analysis, the mean improvement remains between 1.19 and 1.69
points after removing any one dataset, the smallest value occurring when
EuroSAT is omitted.  The complete matched-comparison grid is reported in
\cref{tab:full_grid}; per-method results for each dataset are in
Tables~\ref{tab:per-caltech}--\ref{tab:per-ucf}.

\begin{table*}[t]
\centering
\caption{Top-1 accuracy (\%) at $K=16$, averaged over four backbones. Upper
block: matched implementations under our protocol (\cref{sec:experiments}). Lower block:
published GDA-CLIP figures~\cite{gdaclip}, included as an external consistency check rather than a
matched comparison. The LE-GDA row repeats the numbers above. \textbf{Bold} denotes the
best result per dataset within each block. Means are over 11 datasets in the
upper block and, marked $^{\dagger}$, over 10 in the lower block, where ImageNet
is excluded because our evaluation uses a disjoint 40{,}000-image partition;
the two are therefore not directly comparable.}
\label{tab:per_dataset}
\setlength{\tabcolsep}{4pt}
\footnotesize
\begin{tabular}{lrrrrrrrrrrrr}
\toprule
Method & Caltech & DTD & EuroSAT & Aircraft & Food & ImageNet & Flowers & Pets & Cars & SUN & UCF & Mean\\
\midrule
GDA-CLIP (final-feature) & 93.76 & 69.03 & 84.91 & 42.21 & \textbf{82.51} & 66.47 & 96.08 & \textbf{91.33} & 78.97 & 72.94 & 79.95 & 78.01\\
LE-GDA (ours) & \textbf{94.89} & \textbf{69.71} & \textbf{89.77} & \textbf{46.90} & 82.48 & \textbf{66.63} & \textbf{97.28} & 91.17 & \textbf{79.97} & \textbf{73.84} & \textbf{82.28} & 79.54\\
\addlinespace[2pt]
$\Delta$ (LE-GDA) & +1.13 & +0.68 & +4.86 & +4.69 & $-$0.03 & +0.16 & +1.20 & $-$0.16 & +1.00 & +0.90 & +2.33 & +1.52\\
\midrule
GDA-CLIP (published) & 94.10 & 68.51 & 87.23 & 43.82 & \textbf{82.58} & -- & 96.49 & \textbf{91.30} & 78.82 & 73.06 & 80.28 & 79.62$^{\dagger}$\\
LE-GDA (ours) & \textbf{94.89} & \textbf{69.71} & \textbf{89.77} & \textbf{46.90} & 82.48 & -- & \textbf{97.28} & 91.17 & \textbf{79.97} & \textbf{73.84} & \textbf{82.28} & 80.83$^{\dagger}$\\
\addlinespace[2pt]
$\Delta$ (LE-GDA) & +0.79 & +1.20 & +2.54 & +3.08 & $-$0.10 & -- & +0.79 & $-$0.13 & +1.15 & +0.78 & +2.00 & +1.21$^{\dagger}$\\
\bottomrule
\end{tabular}
\end{table*}

\subsection{Performance across shot counts}
\label{sec:shots}

\begin{figure}[t]
\centering
\begin{subfigure}[t]{\linewidth}
  \centering
  \includegraphics[width=\linewidth]{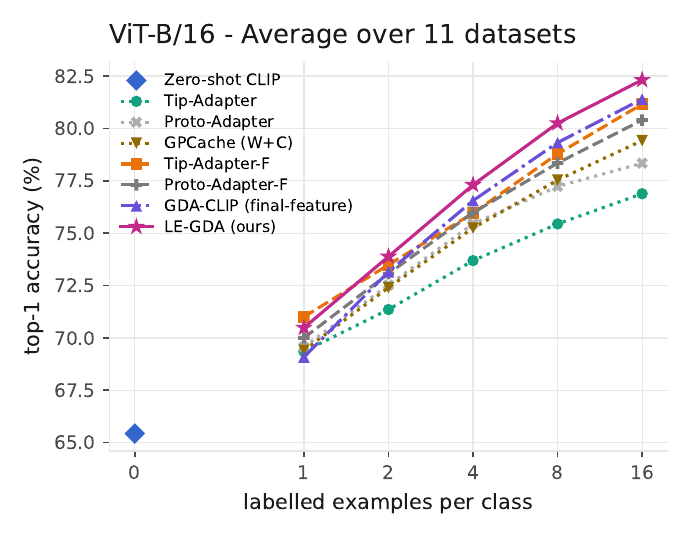}
  \caption{Mean top-1 accuracy across the 11 datasets, ViT-B/16.}
  \label{fig:fewshot-a}
\end{subfigure}

\begin{subfigure}[t]{\linewidth}
  \centering
  \includegraphics[width=\linewidth]{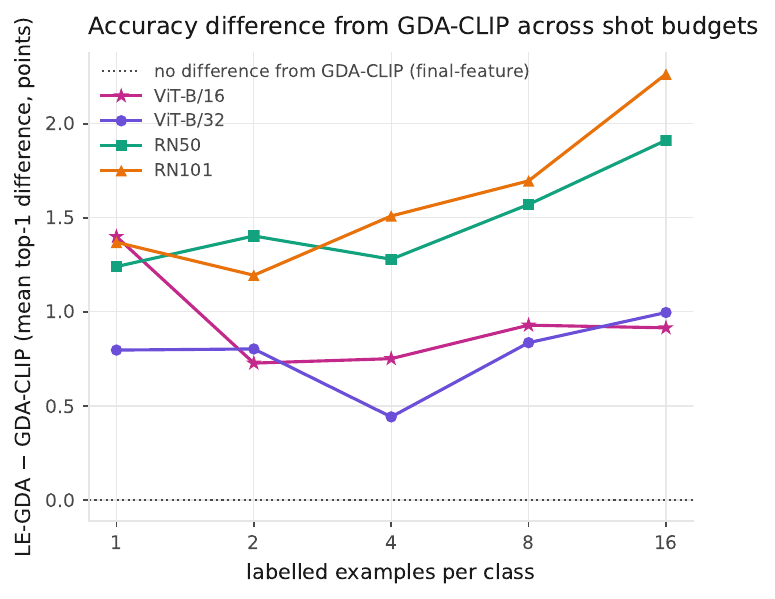}
  \caption{LE-GDA improvement for each CLIP backbone.}
  \label{fig:fewshot-b}
\end{subfigure}
\caption{Few-shot performance as the number of labeled examples per class is
varied.  Results are averaged over 11 datasets in both panels.  The GDA-CLIP
label in the panels denotes the matched GDA-CLIP (final-feature)
implementation described
in \cref{sec:experiments}.  LE-GDA improves this baseline for every
backbone and evaluated shot count (b).}
\label{fig:fewshot}
\end{figure}

\Cref{tab:shots} reports the difference between LE-GDA and the matched
GDA-CLIP (final-feature) baseline.  Every backbone--shot combination is
positive.  The mean improvement over backbones is 1.20, 1.03, 1.00, 1.26, and
1.52 points for $K=1,2,4,8,16$, respectively.  The method therefore improves
accuracy across the full evaluated shot range.
Figure~\ref{fig:fewshot-b} plots these differences.  The gap is positive at
every $K$, and on both ResNet backbones it is largest at $K=16$.

At one shot, every within-class residual is zero, so $\Sigma_\ell=I$ and each
visual classifier is a nearest-class-mean classifier.  Standardization is also
neutral relative to uniform fusion at this shot count (median $-0.003$ points,
\cref{sec:standardization}), yet LE-GDA gains 1.20 points and is positive on
all four backbones.  Together with the matched
search control in \cref{sec:matched_search_control}, this isolates a layer benefit
independent of covariance estimation, standardized fusion, and a larger
final-feature search.

\begin{table}[t]
\centering
\caption{LE-GDA improvement over the matched GDA-CLIP (final-feature) baseline
(accuracy points), averaged over 11 datasets.}
\label{tab:shots}
\small
\resizebox{\columnwidth}{!}{%
\begin{tabular}{lrrrrr}
\toprule
Backbone & $K=1$ & $K=2$ & $K=4$ & $K=8$ & $K=16$\\
\midrule
RN50 &+1.24&+1.40&+1.28&+1.57&+1.91\\
RN101 &+1.37&+1.19&+1.51&+1.70&+2.27\\
ViT-B/32 &+0.80&+0.80&+0.44&+0.84&+1.00\\
ViT-B/16 &+1.40&+0.73&+0.75&+0.93&+0.91\\
\midrule
Mean &+1.20&+1.03&+1.00&+1.26&+1.52\\
\bottomrule
\end{tabular}}
\end{table}

\subsection{Score standardization and representation selection}
\label{sec:standardization}

We compare two fusion rules under otherwise identical conditions. The candidate
representations, ridge and fusion grids, $K$-shot training sets, and greedy
selection procedure are held fixed. Each rule performs its own validation
selection.  The baseline uniformly averages the globally
scaled discriminant scores $\widehat g_\ell$ defined in
\cref{sec:selection_fusion}. LE-GDA applies
the per-query standardization of \cref{eq:zscore} before averaging.

At $K=16$, per-query standardization improves uniform averaging by a median
$0.373$ accuracy points. The comparison covers $11\times4\times3=132$
dataset--backbone--run settings, of which 98 are positive, 31 negative, and
three tied (exact two-sided sign-test $p=2.7\times10^{-9}$).  We also evaluated three alternative fusion rules: within-query score ranks,
the median across standardized spaces, and the mean of softmax probabilities.
All three fall below uniform averaging (\cref{tab:supp-fusion-rules}).  The same median is
$-0.003$ at $K=1$ and $+0.081$ at $K=4$.  This component therefore becomes
more useful as $K$ grows.

Let $d_\ell=\operatorname{median}_{x\in\mathcal V}
\operatorname{std}_c[\widehat g_{\ell,c}(x)]$ denote the validation-set
dispersion of these globally scaled scores under uniform averaging.  This quantity determines the representation's numerical contribution to the
uniform mean. The standard deviation of $\log_{10}d_\ell$ is $0.80$, $1.15$,
$2.35$, and $2.16$ for ViT-B/16, ViT-B/32, RN50, and RN101. The first three
ResNet stages can therefore contribute on the order of $10^{-4}$ of the final
embedding's score variation.  Per-query standardization sets this dispersion to one for every non-degenerate
score vector.  The two rules then select differently.  Under uniform
averaging, selection frequency does not track standalone accuracy on either
ViT backbone.  After per-query standardization, it tracks standalone accuracy
on all four backbones.  The supplementary material reports the per-backbone selection behavior and
two controlled comparisons that separate score dispersion from standalone
accuracy (\cref{sec:supp_selection,tab:supp-dispersion}).
The standardized rule also selects larger multisets: the modal size is
$|\mathcal L^*|=4$, while the median number of distinct spaces is three.
Repetitions therefore reweight mixtures rather than add distinct classifiers.

This ablation characterizes search behavior.  It does not explain LE-GDA's
gain over GDA-CLIP (final-feature).  \Cref{sec:limitations} gives the scope of
the effect.

\subsection{A larger search over one representation}
\label{sec:matched_search_control}

\Cref{tab:search_control} separates the effect of representation diversity
from that of a larger validation search. The matched search control repeats
the same greedy selection and fusion procedure over ridge-regularized versions
of the final embedding, one per candidate space available to LE-GDA on that
backbone (seven for ViTs, five for ResNets), with ridge coefficients spaced
geometrically over $[10^{-6},10^{2}]$. Each control member therefore has a
prescribed ridge coefficient rather than a validation-selected one, so the
control is not an exact nesting of GDA-CLIP (final-feature). It instead tests
whether several ridge variants of one representation produce a systematic
gain. The median difference from GDA-CLIP (final-feature) is zero, whereas
LE-GDA improves the same baseline by 0.829 points. The supplementary material examines error overlap among the selected feature
spaces in \cref{tab:error_overlap}.

Three results constrain the source of the gain. The gain requires multiple representations. It survives at one shot, where neither covariance estimation nor standardized fusion applies. It does not follow from a larger search over one representation. The selected spaces also recover errors that the final embedding makes, although which property of the intermediate features produces the gain
remains open.

\begin{table}[t]
\centering
\caption{Median paired accuracy differences from GDA-CLIP (final-feature) at
$K=16$, over 11 datasets, four backbones, and three seeds.}
\label{tab:search_control}
\small
\begin{tabular}{@{}p{0.78\columnwidth}r@{}}
\toprule
Comparison & Median $\Delta$\\
\midrule
Matched search control vs. GDA-CLIP (final-feature) &+0.000\\
LE-GDA vs. GDA-CLIP (final-feature) &\textbf{+0.829}\\
\bottomrule
\end{tabular}
\end{table}

\subsection{Richer per-class structure within one representation}
\label{sec:mda}

We test whether a single class mean limits LE-GDA. We replace it
with a shared-covariance Gaussian mixture~\cite{mda} using $m$
components per class. The setting is $K=16$ over 10 datasets
(ImageNet excluded), four backbones, and three runs. At $m=1$ the
estimator is identical to LE-GDA. Increasing the number of fixed
components does not help. Relative to $m=1$, mean accuracy falls
monotonically by 0.10, 0.38, 1.10, and 2.10 points at
$m=2,4,8,16$. Selecting $m$ on validation, independently per
feature space, recovers only $+0.020$ points. This is not because
the search stays at $m=1$. It selects $m>1$ in 44\% of individual
space selections and in 87\% of the 120 dataset--backbone--run
settings. The mixture degrees of freedom are therefore available,
and validation cannot convert them into accuracy. The gain comes
from combining representations rather than from richer per-class
structure within one. The ablation protocol and results are provided in
\cref{sec:mda_protocol,tab:mda_results}.

\subsection{Transfer of the selected configuration}
\label{sec:validation_dependence}
We next ask whether LE-GDA's selected configuration transfers
across datasets. We use a leave-one-dataset-out protocol at
$K \in \{4, 16\}$. For each held-out dataset, the representation
multiset, ridge coefficients, and fusion coefficient are derived
from the other ten datasets and their three runs. They are then
applied unchanged. This changes the deployment regime rather than
the label access of the primary comparison. 

Across all 44 dataset--backbone settings at $K = 16$, the paired
difference is $+0.611 \pm 2.771$ points with 28 wins. The
sign-test gives $p = 0.096$ and the Wilcoxon test $p = 0.126$,
both two-sided. Neither test provides strong evidence of a
difference. The advantage is also concentrated on EuroSAT.
Averaging over backbones within each dataset and excluding it,
LE-GDA leads by $0.09$ points over the remaining ten datasets.

Full validation access improves LE-GDA by $2.270 \pm 2.276$
points. The fusion coefficient contributes $1.046$ points, the
representation multiset $0.941$, and the ridge coefficient
$0.283$. The transferred $\alpha$ lies on a shallow validation
plateau. It nonetheless accounts for 46\% of the difference. A
flat validation curve therefore does not establish that a
hyperparameter transfers without loss. The supplementary material gives the intermediate configurations,
per-component decomposition, and $K=4$ results in
\cref{tab:transfer_configs,tab:transfer_contrasts,tab:transfer_k4}.

\begin{table}[t]
\centering
\caption{Decomposition of the $2.270$-point difference between fully
validation-selected and fully transferred LE-GDA at $K=16$.  Differences are
paired over 11 datasets and four backbones.}
\label{tab:validation_decomposition}
\small
\begin{tabular*}{\columnwidth}{@{\extracolsep{\fill}}lrr}
\toprule
Selected component & Mean $\Delta\pm$ sd & Share\\
\midrule
Feature-space multiset &$+0.941\pm1.269$&41\%\\
Fusion coefficient $\alpha$ &$+1.046\pm1.487$&46\%\\
Ridge coefficient $\lambda$ &$+0.283\pm0.875$&$\sim$13\%\\
\bottomrule
\end{tabular*}
\end{table}

\FloatBarrier
\section{Limitations}
\label{sec:limitations}

Tip-Adapter and Tip-Adapter-F are run from the authors' released code with
their own hyperparameter grids. Our GPCache and Proto-Adapter implementations
use finite validation grids for their affinity scale and fusion coefficients.
These comparisons therefore do not establish the best performance attainable
by every cache formulation. ImageNet is evaluated on our disjoint
40,000-image partition (Sec.~\ref{sec:experiments}), where LE-GDA underperforms both gradient-trained cache
adapters on every backbone. Averaged over backbones, it has the highest
ImageNet accuracy among training-free methods.

The dispersion analysis characterizes which representations each fusion rule
selects.  It does not generally predict where standardization improves
accuracy.  Dispersion heterogeneity is associated with the accuracy gain on
RN50 and RN101, but not on either ViT backbone.  On ViT-B/16 the 16-shot
accuracy difference is 0.05 points despite a large change in the selected
representations.  The selection analysis is moreover conducted only at $K=16$.

The validation images are additional labeled data outside the $K$-shot
training set.  Our primary results therefore characterize validation-assisted
few-shot adaptation.  When all selected quantities are transferred by
leave-one-dataset-out, validation-free LE-GDA does not significantly improve
the validation-free baseline.  

LE-GDA has no trainable parameters and requires no backward pass. Fitting
reduces to closed-form statistics and a validation search over three discrete
choices. It stores three to five times more classifier values than GDA-CLIP (final-feature). That remains only 19--31\% of a sparse 16-shot Tip-Adapter cache. The supplementary material reports the per-backbone accounting in \cref{tab:storage}. It also requires access to intermediate activations, so it cannot be applied when only the final embedding is exposed.
Finally, we leave distribution shift and base-to-new generalization to future work.
\section{Conclusion}

We asked whether intermediate visual representations can improve training-free
few-shot adaptation.  LE-GDA fits regularized Gaussian discriminants to
intermediate and final CLIP features. It standardizes their
class scores per query, and fuses a validation-selected subset with the
zero-shot classifier.  It outperforms a matched GDA-CLIP (final-feature)
baseline for every evaluated backbone and shot count, including at one shot,
where neither covariance estimation nor standardized fusion explains the gain.
Intermediate CLIP representations thus contain useful information left unused
by final-feature methods.  The results also identify layer choice, score normalization, and
matched validation access as consequential design decisions.

{\small
\bibliographystyle{ieeenat_fullname}
\bibliography{references}
}

\clearpage\appendix
\renewcommand{\thetable}{S\arabic{table}}
\renewcommand{\thefigure}{S\arabic{figure}}
\renewcommand{\theequation}{S\arabic{equation}}
\renewcommand{\theHtable}{S\arabic{table}}
\renewcommand{\theHfigure}{S\arabic{figure}}
\renewcommand{\theHequation}{S\arabic{equation}}
\setcounter{table}{0}\setcounter{figure}{0}\setcounter{equation}{0}

\begin{center}
{\large\bf Supplementary Material}
\end{center}
\vspace{0.5em}

\newcommand{\bestcell}[2]{{\bfseries #1{\scriptsize\boldmath$\pm$#2}}}

\section{Complete dataset--backbone results}

All ImageNet accuracies follow the fixed, class-balanced validation/evaluation
partition defined in the main paper.

Throughout this supplementary material, GDA-CLIP (final-feature)
denotes our matched reimplementation evaluated under the same data splits,
features, and model-selection protocol as LE-GDA.  Published GDA-CLIP
refers to the original method, estimator, or numerical results reported
in~\cite{gdaclip}.

\Cref{tab:full_grid} reports the complete comparison between LE-GDA and
the matched GDA-CLIP (final-feature) baseline at $K=16$.  Each entry is averaged over
the three independently sampled $K$-shot training sets used in the main paper.

\begin{table*}[t]
\centering
\caption{Top-1 accuracy (\%) for every dataset and CLIP backbone at
$K=16$, for LE-GDA and its matched baseline only. Results are averaged over
three independently sampled $K$-shot training sets.
Tables~\ref{tab:per-caltech}--\ref{tab:per-ucf} give the same settings for
all eight methods.}
\label{tab:full_grid}
\small
\resizebox{\textwidth}{!}{%
\begin{tabular}{llrrrrrrrrrrrr}
\toprule
Backbone & Method & Caltech & DTD & EuroSAT & Aircraft & Food & ImageNet &
Flowers & Pets & Cars & SUN & UCF & Mean\\
\midrule
RN50 & GDA-CLIP (final-feature) &91.93&67.18&83.49&39.06&79.02&62.37&95.44&89.53&74.83&70.45&77.86&75.56\\
 & LE-GDA &\textbf{93.67}&\textbf{68.18}&\textbf{88.97}&\textbf{44.45}&\textbf{79.07}&\textbf{62.82}&\textbf{97.09}&89.23&\textbf{76.55}&\textbf{71.49}&\textbf{80.67}&\textbf{77.47}\\
\midrule
RN101 & GDA-CLIP (final-feature) &93.74&68.64&83.02&40.23&82.10&65.41&95.57&\textbf{91.24}&80.52&72.10&78.94&77.41\\
 & LE-GDA &\textbf{95.09}&\textbf{70.39}&\textbf{89.22}&\textbf{48.06}&\textbf{82.15}&\textbf{65.88}&\textbf{97.77}&90.95&\textbf{82.03}&\textbf{73.22}&\textbf{81.67}&\textbf{79.68}\\
\midrule
ViT-B/32 & GDA-CLIP (final-feature) &94.27&67.79&85.62&40.64&\textbf{81.76}&\textbf{66.67}&95.95&90.82&77.51&73.46&80.14&77.69\\
 & LE-GDA &\textbf{94.97}&\textbf{68.16}&\textbf{90.34}&\textbf{43.15}&81.58&66.46&\textbf{96.35}&\textbf{90.86}&\textbf{77.94}&\textbf{74.14}&\textbf{81.66}&\textbf{78.69}\\
\midrule
ViT-B/16 & GDA-CLIP (final-feature) &95.10&\textbf{72.52}&87.53&48.91&\textbf{87.16}&\textbf{71.44}&97.36&\textbf{93.73}&83.01&75.77&82.84&81.40\\
 & LE-GDA &\textbf{95.84}&72.12&\textbf{90.55}&\textbf{51.94}&87.12&71.36&\textbf{97.90}&93.66&\textbf{83.36}&\textbf{76.50}&\textbf{85.10}&\textbf{82.31}\\
\bottomrule
\end{tabular}}
\end{table*}

Across the $11\times4=44$ dataset--backbone comparisons in
\cref{tab:full_grid}, LE-GDA improves GDA-CLIP (final-feature) in 36 cases.
The largest gains occur on EuroSAT and FGVC-Aircraft, including gains of 6.20
and 7.83 points with RN101.  The eight negative differences are small: the
largest is 0.40 points.  The complete grid therefore supports the main-paper
finding that the improvement is broad across backbones and is not produced by
a single dataset or architecture.

\subsection{Per-method comparisons by dataset and backbone}
\label{sec:per_dataset_methods}

The following tables retain the dataset axis that is averaged out in the
main paper's aggregate backbone comparison.  Every backbone entry is the
mean top-1 accuracy (\%) with its standard deviation over the three
independently sampled $K=16$ training sets; the final column averages the
four backbone means.  Bold denotes the best adaptation method in each
backbone column, excluding zero-shot CLIP.  The Grad. column marks the two
methods that use gradient-based adaptation for 20 epochs.  The Impl. column
identifies results produced by the original authors' released code (orig.)
or by our implementation (ours), always under our protocol.  Tip-Adapter and
Tip-Adapter-F use the original implementation; all other adaptation methods
use ours.  Tables~\ref{tab:per-caltech}--\ref{tab:per-ucf} report one dataset
each, in the order used in the main paper's Table~2.

\begin{table*}[t]
\centering
\caption{Per-method top-1 accuracy (\%) on Caltech101 at $K=16$.}
\label{tab:per-caltech}
\small
\resizebox{\textwidth}{!}{%
\begin{tabular}{lccrrrrr}
\toprule
Method & Grad. & Impl. & RN50 & RN101 & ViT-B/32 & ViT-B/16 & Mean\\
\midrule
Zero-shot CLIP & -- & — & 85.72{\scriptsize$\pm$.00} & 89.98{\scriptsize$\pm$.00} & 91.48{\scriptsize$\pm$.00} & 93.27{\scriptsize$\pm$.00} & 90.11\\
Tip-Adapter & -- & orig. & 90.33{\scriptsize$\pm$.12} & 92.95{\scriptsize$\pm$.13} & 93.81{\scriptsize$\pm$.37} & 94.86{\scriptsize$\pm$.19} & 92.99\\
Proto-Adapter & -- & ours & 91.63{\scriptsize$\pm$.07} & 93.37{\scriptsize$\pm$.22} & 93.78{\scriptsize$\pm$.30} & 95.36{\scriptsize$\pm$.23} & 93.54\\
GPCache (W+C) & -- & ours & 91.62{\scriptsize$\pm$.12} & 93.63{\scriptsize$\pm$.22} & 94.37{\scriptsize$\pm$.18} & 95.36{\scriptsize$\pm$.14} & 93.75\\
Tip-Adapter-F & \checkmark & orig. & 92.45{\scriptsize$\pm$.21} & 94.55{\scriptsize$\pm$.05} & \bestcell{95.29}{.22} & 95.81{\scriptsize$\pm$.11} & 94.53\\
Proto-Adapter-F & \checkmark & ours & 92.52{\scriptsize$\pm$.23} & 94.16{\scriptsize$\pm$.40} & 94.86{\scriptsize$\pm$.16} & 95.66{\scriptsize$\pm$.43} & 94.30\\
GDA-CLIP (final-feature) & -- & ours & 91.93{\scriptsize$\pm$.56} & 93.74{\scriptsize$\pm$.46} & 94.27{\scriptsize$\pm$.36} & 95.10{\scriptsize$\pm$.50} & 93.76\\
LE-GDA (ours) & -- & ours & \bestcell{93.67}{.39} & \bestcell{95.09}{.22} & 94.97{\scriptsize$\pm$.37} & \bestcell{95.84}{.40} & 94.89\\
\bottomrule
\end{tabular}}
\end{table*}
\begin{table*}[t]
\centering
\caption{Per-method top-1 accuracy (\%) on DTD at $K=16$.}
\label{tab:per-dtd}
\small
\resizebox{\textwidth}{!}{%
\begin{tabular}{lccrrrrr}
\toprule
Method & Grad. & Impl. & RN50 & RN101 & ViT-B/32 & ViT-B/16 & Mean\\
\midrule
Zero-shot CLIP & -- & — & 42.97{\scriptsize$\pm$.00} & 38.12{\scriptsize$\pm$.00} & 44.15{\scriptsize$\pm$.00} & 44.03{\scriptsize$\pm$.00} & 42.32\\
Tip-Adapter & -- & orig. & 60.66{\scriptsize$\pm$.17} & 63.04{\scriptsize$\pm$.36} & 62.61{\scriptsize$\pm$.42} & 65.54{\scriptsize$\pm$.39} & 62.96\\
Proto-Adapter & -- & ours & 65.01{\scriptsize$\pm$.34} & 64.81{\scriptsize$\pm$.44} & 65.35{\scriptsize$\pm$.41} & 67.45{\scriptsize$\pm$.74} & 65.66\\
GPCache (W+C) & -- & ours & 64.76{\scriptsize$\pm$.62} & 67.42{\scriptsize$\pm$.70} & 66.35{\scriptsize$\pm$.64} & 69.94{\scriptsize$\pm$.28} & 67.11\\
Tip-Adapter-F & \checkmark & orig. & 66.21{\scriptsize$\pm$.23} & 69.29{\scriptsize$\pm$.06} & \bestcell{69.13}{.84} & 72.16{\scriptsize$\pm$.48} & 69.20\\
Proto-Adapter-F & \checkmark & ours & 67.63{\scriptsize$\pm$.78} & 68.09{\scriptsize$\pm$.60} & 67.34{\scriptsize$\pm$.52} & 70.41{\scriptsize$\pm$.66} & 68.37\\
GDA-CLIP (final-feature) & -- & ours & 67.18{\scriptsize$\pm$.48} & 68.64{\scriptsize$\pm$.07} & 67.79{\scriptsize$\pm$.32} & \bestcell{72.52}{.48} & 69.03\\
LE-GDA (ours) & -- & ours & \bestcell{68.18}{.24} & \bestcell{70.39}{.99} & 68.16{\scriptsize$\pm$.24} & 72.12{\scriptsize$\pm$1.11} & 69.72\\
\bottomrule
\end{tabular}}
\end{table*}
\begin{table*}[t]
\centering
\caption{Per-method top-1 accuracy (\%) on EuroSAT at $K=16$.}
\label{tab:per-eurosat}
\small
\resizebox{\textwidth}{!}{%
\begin{tabular}{lccrrrrr}
\toprule
Method & Grad. & Impl. & RN50 & RN101 & ViT-B/32 & ViT-B/16 & Mean\\
\midrule
Zero-shot CLIP & -- & — & 36.27{\scriptsize$\pm$.00} & 32.78{\scriptsize$\pm$.00} & 45.31{\scriptsize$\pm$.00} & 48.49{\scriptsize$\pm$.00} & 40.71\\
Tip-Adapter & -- & orig. & 71.67{\scriptsize$\pm$1.24} & 68.47{\scriptsize$\pm$1.90} & 74.14{\scriptsize$\pm$.90} & 77.87{\scriptsize$\pm$.35} & 73.04\\
Proto-Adapter & -- & ours & 74.84{\scriptsize$\pm$1.10} & 69.22{\scriptsize$\pm$1.20} & 75.93{\scriptsize$\pm$.54} & 79.10{\scriptsize$\pm$.34} & 74.77\\
GPCache (W+C) & -- & ours & 77.60{\scriptsize$\pm$1.11} & 76.31{\scriptsize$\pm$.49} & 80.76{\scriptsize$\pm$.25} & 83.26{\scriptsize$\pm$.70} & 79.48\\
Tip-Adapter-F & \checkmark & orig. & 85.28{\scriptsize$\pm$.82} & 82.89{\scriptsize$\pm$.90} & 85.87{\scriptsize$\pm$.15} & 87.56{\scriptsize$\pm$.08} & 85.40\\
Proto-Adapter-F & \checkmark & ours & 80.08{\scriptsize$\pm$.83} & 74.66{\scriptsize$\pm$.96} & 79.78{\scriptsize$\pm$.47} & 81.15{\scriptsize$\pm$.82} & 78.92\\
GDA-CLIP (final-feature) & -- & ours & 83.49{\scriptsize$\pm$.56} & 83.02{\scriptsize$\pm$.20} & 85.62{\scriptsize$\pm$.31} & 87.53{\scriptsize$\pm$.10} & 84.92\\
LE-GDA (ours) & -- & ours & \bestcell{88.97}{.36} & \bestcell{89.22}{.95} & \bestcell{90.34}{.54} & \bestcell{90.55}{.06} & 89.77\\
\bottomrule
\end{tabular}}
\end{table*}
\begin{table*}[t]
\centering
\caption{Per-method top-1 accuracy (\%) on FGVC-Aircraft at $K=16$.}
\label{tab:per-aircraft}
\small
\resizebox{\textwidth}{!}{%
\begin{tabular}{lccrrrrr}
\toprule
Method & Grad. & Impl. & RN50 & RN101 & ViT-B/32 & ViT-B/16 & Mean\\
\midrule
Zero-shot CLIP & -- & — & 17.10{\scriptsize$\pm$.00} & 17.91{\scriptsize$\pm$.00} & 19.08{\scriptsize$\pm$.00} & 24.81{\scriptsize$\pm$.00} & 19.73\\
Tip-Adapter & -- & orig. & 29.61{\scriptsize$\pm$.43} & 31.73{\scriptsize$\pm$.40} & 31.41{\scriptsize$\pm$.89} & 40.46{\scriptsize$\pm$.48} & 33.31\\
Proto-Adapter & -- & ours & 27.63{\scriptsize$\pm$.28} & 29.01{\scriptsize$\pm$.36} & 31.17{\scriptsize$\pm$.30} & 38.83{\scriptsize$\pm$.57} & 31.66\\
GPCache (W+C) & -- & ours & 32.48{\scriptsize$\pm$.20} & 35.96{\scriptsize$\pm$.81} & 35.48{\scriptsize$\pm$.09} & 43.91{\scriptsize$\pm$.37} & 36.96\\
Tip-Adapter-F & \checkmark & orig. & 35.40{\scriptsize$\pm$.37} & 37.66{\scriptsize$\pm$.67} & 36.69{\scriptsize$\pm$.21} & 45.75{\scriptsize$\pm$.32} & 38.88\\
Proto-Adapter-F & \checkmark & ours & 35.79{\scriptsize$\pm$.17} & 37.55{\scriptsize$\pm$.22} & 36.87{\scriptsize$\pm$.38} & 45.42{\scriptsize$\pm$.51} & 38.91\\
GDA-CLIP (final-feature) & -- & ours & 39.06{\scriptsize$\pm$.41} & 40.23{\scriptsize$\pm$.99} & 40.64{\scriptsize$\pm$.36} & 48.91{\scriptsize$\pm$.40} & 42.21\\
LE-GDA (ours) & -- & ours & \bestcell{44.45}{.46} & \bestcell{48.06}{.45} & \bestcell{43.15}{.26} & \bestcell{51.94}{.32} & 46.90\\
\bottomrule
\end{tabular}}
\end{table*}
\begin{table*}[t]
\centering
\caption{Per-method top-1 accuracy (\%) on Food101 at $K=16$.}
\label{tab:per-food}
\small
\resizebox{\textwidth}{!}{%
\begin{tabular}{lccrrrrr}
\toprule
Method & Grad. & Impl. & RN50 & RN101 & ViT-B/32 & ViT-B/16 & Mean\\
\midrule
Zero-shot CLIP & -- & — & 77.38{\scriptsize$\pm$.00} & 80.68{\scriptsize$\pm$.00} & 80.41{\scriptsize$\pm$.00} & 85.88{\scriptsize$\pm$.00} & 81.09\\
Tip-Adapter & -- & orig. & 77.85{\scriptsize$\pm$.13} & 81.05{\scriptsize$\pm$.06} & 80.89{\scriptsize$\pm$.05} & 86.36{\scriptsize$\pm$.12} & 81.54\\
Proto-Adapter & -- & ours & 78.62{\scriptsize$\pm$.06} & 81.74{\scriptsize$\pm$.10} & 81.25{\scriptsize$\pm$.10} & 86.93{\scriptsize$\pm$.04} & 82.13\\
GPCache (W+C) & -- & ours & 78.29{\scriptsize$\pm$.10} & 81.73{\scriptsize$\pm$.09} & 81.30{\scriptsize$\pm$.09} & 86.94{\scriptsize$\pm$.04} & 82.06\\
Tip-Adapter-F & \checkmark & orig. & \bestcell{79.33}{.14} & \bestcell{82.22}{.26} & \bestcell{82.16}{.06} & \bestcell{87.30}{.07} & 82.75\\
Proto-Adapter-F & \checkmark & ours & 79.01{\scriptsize$\pm$.04} & 82.10{\scriptsize$\pm$.12} & 81.64{\scriptsize$\pm$.11} & 87.12{\scriptsize$\pm$.15} & 82.47\\
GDA-CLIP (final-feature) & -- & ours & 79.02{\scriptsize$\pm$.04} & 82.10{\scriptsize$\pm$.05} & 81.76{\scriptsize$\pm$.11} & 87.16{\scriptsize$\pm$.12} & 82.51\\
LE-GDA (ours) & -- & ours & 79.07{\scriptsize$\pm$.15} & 82.15{\scriptsize$\pm$.21} & 81.58{\scriptsize$\pm$.16} & 87.12{\scriptsize$\pm$.05} & 82.48\\
\bottomrule
\end{tabular}}
\end{table*}
\begin{table*}[t]
\centering
\caption{Per-method top-1 accuracy (\%) on ImageNet at $K=16$.}
\label{tab:per-imagenet}
\small
\resizebox{\textwidth}{!}{%
\begin{tabular}{lccrrrrr}
\toprule
Method & Grad. & Impl. & RN50 & RN101 & ViT-B/32 & ViT-B/16 & Mean\\
\midrule
Zero-shot CLIP & -- & — & 59.15{\scriptsize$\pm$.00} & 61.25{\scriptsize$\pm$.00} & 63.13{\scriptsize$\pm$.00} & 67.84{\scriptsize$\pm$.00} & 62.84\\
Tip-Adapter & -- & orig. & 60.39{\scriptsize$\pm$.06} & 63.10{\scriptsize$\pm$.11} & 64.70{\scriptsize$\pm$.05} & 69.47{\scriptsize$\pm$.08} & 64.42\\
Proto-Adapter & -- & ours & 61.97{\scriptsize$\pm$.09} & 64.85{\scriptsize$\pm$.16} & 65.90{\scriptsize$\pm$.05} & 70.77{\scriptsize$\pm$.08} & 65.87\\
GPCache (W+C) & -- & ours & 61.50{\scriptsize$\pm$.10} & 64.47{\scriptsize$\pm$.05} & 65.73{\scriptsize$\pm$.10} & 70.58{\scriptsize$\pm$.09} & 65.57\\
Tip-Adapter-F & \checkmark & orig. & 63.29{\scriptsize$\pm$.03} & \bestcell{66.36}{.03} & 67.33{\scriptsize$\pm$.07} & 71.98{\scriptsize$\pm$.07} & 67.24\\
Proto-Adapter-F & \checkmark & ours & \bestcell{63.42}{.03} & 66.28{\scriptsize$\pm$.08} & \bestcell{67.49}{.08} & \bestcell{72.19}{.05} & 67.35\\
GDA-CLIP (final-feature) & -- & ours & 62.37{\scriptsize$\pm$.09} & 65.41{\scriptsize$\pm$.04} & 66.67{\scriptsize$\pm$.20} & 71.44{\scriptsize$\pm$.19} & 66.47\\
LE-GDA (ours) & -- & ours & 62.82{\scriptsize$\pm$.15} & 65.88{\scriptsize$\pm$.17} & 66.46{\scriptsize$\pm$.14} & 71.36{\scriptsize$\pm$.23} & 66.63\\
\bottomrule
\end{tabular}}
\end{table*}
\begin{table*}[t]
\centering
\caption{Per-method top-1 accuracy (\%) on Flowers102 at $K=16$.}
\label{tab:per-flowers}
\small
\resizebox{\textwidth}{!}{%
\begin{tabular}{lccrrrrr}
\toprule
Method & Grad. & Impl. & RN50 & RN101 & ViT-B/32 & ViT-B/16 & Mean\\
\midrule
Zero-shot CLIP & -- & — & 66.06{\scriptsize$\pm$.00} & 64.43{\scriptsize$\pm$.00} & 66.50{\scriptsize$\pm$.00} & 70.81{\scriptsize$\pm$.00} & 66.95\\
Tip-Adapter & -- & orig. & 89.39{\scriptsize$\pm$.32} & 90.19{\scriptsize$\pm$.30} & 91.18{\scriptsize$\pm$.07} & 93.90{\scriptsize$\pm$.07} & 91.16\\
Proto-Adapter & -- & ours & 92.11{\scriptsize$\pm$.18} & 92.34{\scriptsize$\pm$.15} & 92.66{\scriptsize$\pm$.13} & 95.60{\scriptsize$\pm$.08} & 93.18\\
GPCache (W+C) & -- & ours & 92.72{\scriptsize$\pm$.12} & 94.68{\scriptsize$\pm$.17} & 94.30{\scriptsize$\pm$.38} & 96.91{\scriptsize$\pm$.29} & 94.65\\
Tip-Adapter-F & \checkmark & orig. & 94.56{\scriptsize$\pm$.58} & 94.92{\scriptsize$\pm$.10} & 93.99{\scriptsize$\pm$.26} & 96.52{\scriptsize$\pm$.32} & 95.00\\
Proto-Adapter-F & \checkmark & ours & 95.55{\scriptsize$\pm$.05} & 95.82{\scriptsize$\pm$.25} & 95.28{\scriptsize$\pm$.02} & 97.28{\scriptsize$\pm$.03} & 95.98\\
GDA-CLIP (final-feature) & -- & ours & 95.44{\scriptsize$\pm$.20} & 95.57{\scriptsize$\pm$.33} & 95.95{\scriptsize$\pm$.10} & 97.36{\scriptsize$\pm$.18} & 96.08\\
LE-GDA (ours) & -- & ours & \bestcell{97.09}{.21} & \bestcell{97.77}{.03} & \bestcell{96.35}{.14} & \bestcell{97.90}{.05} & 97.28\\
\bottomrule
\end{tabular}}
\end{table*}
\begin{table*}[t]
\centering
\caption{Per-method top-1 accuracy (\%) on OxfordPets at $K=16$.}
\label{tab:per-pets}
\small
\resizebox{\textwidth}{!}{%
\begin{tabular}{lccrrrrr}
\toprule
Method & Grad. & Impl. & RN50 & RN101 & ViT-B/32 & ViT-B/16 & Mean\\
\midrule
Zero-shot CLIP & -- & — & 85.80{\scriptsize$\pm$.00} & 86.94{\scriptsize$\pm$.00} & 87.44{\scriptsize$\pm$.00} & 89.10{\scriptsize$\pm$.00} & 87.32\\
Tip-Adapter & -- & orig. & 87.90{\scriptsize$\pm$.25} & 87.65{\scriptsize$\pm$.10} & 88.27{\scriptsize$\pm$.57} & 91.74{\scriptsize$\pm$.50} & 88.89\\
Proto-Adapter & -- & ours & 88.23{\scriptsize$\pm$.19} & 89.21{\scriptsize$\pm$.40} & 89.56{\scriptsize$\pm$.21} & 92.15{\scriptsize$\pm$.17} & 89.79\\
GPCache (W+C) & -- & ours & 87.97{\scriptsize$\pm$.06} & 88.90{\scriptsize$\pm$.53} & 89.32{\scriptsize$\pm$.75} & 92.40{\scriptsize$\pm$.34} & 89.65\\
Tip-Adapter-F & \checkmark & orig. & 89.30{\scriptsize$\pm$.19} & 90.39{\scriptsize$\pm$.22} & 90.72{\scriptsize$\pm$.32} & 92.96{\scriptsize$\pm$.29} & 90.84\\
Proto-Adapter-F & \checkmark & ours & 88.94{\scriptsize$\pm$.45} & 90.88{\scriptsize$\pm$.32} & 89.73{\scriptsize$\pm$.08} & 92.80{\scriptsize$\pm$.23} & 90.59\\
GDA-CLIP (final-feature) & -- & ours & \bestcell{89.53}{.18} & \bestcell{91.24}{.73} & 90.82{\scriptsize$\pm$.17} & \bestcell{93.73}{.12} & 91.33\\
LE-GDA (ours) & -- & ours & 89.23{\scriptsize$\pm$.61} & 90.95{\scriptsize$\pm$.44} & \bestcell{90.86}{.15} & 93.66{\scriptsize$\pm$.13} & 91.18\\
\bottomrule
\end{tabular}}
\end{table*}
\begin{table*}[t]
\centering
\caption{Per-method top-1 accuracy (\%) on StanfordCars at $K=16$.}
\label{tab:per-cars}
\small
\resizebox{\textwidth}{!}{%
\begin{tabular}{lccrrrrr}
\toprule
Method & Grad. & Impl. & RN50 & RN101 & ViT-B/32 & ViT-B/16 & Mean\\
\midrule
Zero-shot CLIP & -- & — & 55.73{\scriptsize$\pm$.00} & 63.21{\scriptsize$\pm$.00} & 60.20{\scriptsize$\pm$.00} & 65.55{\scriptsize$\pm$.00} & 61.17\\
Tip-Adapter & -- & orig. & 66.14{\scriptsize$\pm$.39} & 72.71{\scriptsize$\pm$.18} & 70.01{\scriptsize$\pm$.33} & 75.52{\scriptsize$\pm$.31} & 71.10\\
Proto-Adapter & -- & ours & 70.32{\scriptsize$\pm$.19} & 76.24{\scriptsize$\pm$.56} & 72.81{\scriptsize$\pm$.11} & 79.20{\scriptsize$\pm$.29} & 74.64\\
GPCache (W+C) & -- & ours & 69.59{\scriptsize$\pm$.60} & 76.79{\scriptsize$\pm$.39} & 73.25{\scriptsize$\pm$.12} & 78.59{\scriptsize$\pm$.11} & 74.56\\
Tip-Adapter-F & \checkmark & orig. & 74.92{\scriptsize$\pm$.56} & 80.67{\scriptsize$\pm$.39} & 77.88{\scriptsize$\pm$.59} & 82.76{\scriptsize$\pm$.06} & 79.06\\
Proto-Adapter-F & \checkmark & ours & 75.97{\scriptsize$\pm$.51} & 80.81{\scriptsize$\pm$.41} & 77.10{\scriptsize$\pm$.40} & 82.78{\scriptsize$\pm$.48} & 79.16\\
GDA-CLIP (final-feature) & -- & ours & 74.83{\scriptsize$\pm$.42} & 80.52{\scriptsize$\pm$.46} & 77.51{\scriptsize$\pm$.66} & 83.01{\scriptsize$\pm$.32} & 78.97\\
LE-GDA (ours) & -- & ours & \bestcell{76.55}{.34} & \bestcell{82.03}{.51} & \bestcell{77.94}{.36} & \bestcell{83.36}{.35} & 79.97\\
\bottomrule
\end{tabular}}
\end{table*}
\begin{table*}[t]
\centering
\caption{Per-method top-1 accuracy (\%) on SUN397 at $K=16$.}
\label{tab:per-sun}
\small
\resizebox{\textwidth}{!}{%
\begin{tabular}{lccrrrrr}
\toprule
Method & Grad. & Impl. & RN50 & RN101 & ViT-B/32 & ViT-B/16 & Mean\\
\midrule
Zero-shot CLIP & -- & — & 58.85{\scriptsize$\pm$.00} & 59.02{\scriptsize$\pm$.00} & 62.06{\scriptsize$\pm$.00} & 62.56{\scriptsize$\pm$.00} & 60.62\\
Tip-Adapter & -- & orig. & 67.02{\scriptsize$\pm$.24} & 68.19{\scriptsize$\pm$.19} & 70.26{\scriptsize$\pm$.22} & 71.71{\scriptsize$\pm$.22} & 69.30\\
Proto-Adapter & -- & ours & 70.09{\scriptsize$\pm$.15} & 71.05{\scriptsize$\pm$.15} & 72.80{\scriptsize$\pm$.06} & 74.62{\scriptsize$\pm$.30} & 72.14\\
GPCache (W+C) & -- & ours & 69.24{\scriptsize$\pm$.14} & 70.99{\scriptsize$\pm$.20} & 72.63{\scriptsize$\pm$.08} & 74.38{\scriptsize$\pm$.36} & 71.81\\
Tip-Adapter-F & \checkmark & orig. & 71.29{\scriptsize$\pm$.05} & 72.83{\scriptsize$\pm$.17} & 74.02{\scriptsize$\pm$.22} & 76.17{\scriptsize$\pm$.16} & 73.58\\
Proto-Adapter-F & \checkmark & ours & \bestcell{71.61}{.13} & 72.94{\scriptsize$\pm$.09} & 74.12{\scriptsize$\pm$.14} & 76.16{\scriptsize$\pm$.25} & 73.71\\
GDA-CLIP (final-feature) & -- & ours & 70.45{\scriptsize$\pm$.40} & 72.10{\scriptsize$\pm$.16} & 73.46{\scriptsize$\pm$.07} & 75.77{\scriptsize$\pm$.15} & 72.95\\
LE-GDA (ours) & -- & ours & 71.49{\scriptsize$\pm$.07} & \bestcell{73.22}{.21} & \bestcell{74.14}{.13} & \bestcell{76.50}{.10} & 73.84\\
\bottomrule
\end{tabular}}
\end{table*}
\begin{table*}[t]
\centering
\caption{Per-method top-1 accuracy (\%) on UCF101 at $K=16$.}
\label{tab:per-ucf}
\small
\resizebox{\textwidth}{!}{%
\begin{tabular}{lccrrrrr}
\toprule
Method & Grad. & Impl. & RN50 & RN101 & ViT-B/32 & ViT-B/16 & Mean\\
\midrule
Zero-shot CLIP & -- & — & 61.86{\scriptsize$\pm$.00} & 61.12{\scriptsize$\pm$.00} & 63.49{\scriptsize$\pm$.00} & 67.41{\scriptsize$\pm$.00} & 63.47\\
Tip-Adapter & -- & orig. & 71.27{\scriptsize$\pm$.24} & 73.52{\scriptsize$\pm$.33} & 74.39{\scriptsize$\pm$.30} & 78.26{\scriptsize$\pm$.22} & 74.36\\
Proto-Adapter & -- & ours & 76.80{\scriptsize$\pm$.95} & 78.20{\scriptsize$\pm$.28} & 78.46{\scriptsize$\pm$.43} & 81.69{\scriptsize$\pm$.23} & 78.79\\
GPCache (W+C) & -- & ours & 75.12{\scriptsize$\pm$.11} & 77.01{\scriptsize$\pm$.31} & 78.21{\scriptsize$\pm$.25} & 81.32{\scriptsize$\pm$.22} & 77.91\\
Tip-Adapter-F & \checkmark & orig. & 78.52{\scriptsize$\pm$.79} & 80.66{\scriptsize$\pm$.35} & 80.66{\scriptsize$\pm$.54} & 83.79{\scriptsize$\pm$.17} & 80.91\\
Proto-Adapter-F & \checkmark & ours & 78.76{\scriptsize$\pm$.84} & 80.08{\scriptsize$\pm$.36} & 80.17{\scriptsize$\pm$.30} & 83.35{\scriptsize$\pm$.08} & 80.59\\
GDA-CLIP (final-feature) & -- & ours & 77.86{\scriptsize$\pm$.90} & 78.94{\scriptsize$\pm$.50} & 80.14{\scriptsize$\pm$.46} & 82.84{\scriptsize$\pm$.30} & 79.95\\
LE-GDA (ours) & -- & ours & \bestcell{80.67}{.20} & \bestcell{81.67}{.21} & \bestcell{81.66}{.03} & \bestcell{85.10}{.09} & 82.28\\
\bottomrule
\end{tabular}}
\end{table*}

\FloatBarrier

\begin{strip}
\begin{minipage}[t]{0.56\textwidth}
\centering
\captionsetup{hypcap=false}
\captionof{table}{Published results versus our matched reimplementations at $K=16$,
averaged over 11 datasets. Published GDA-CLIP, Tip-Adapter and Tip-Adapter-F
values are from~\cite{gdaclip}; published Proto-Adapter values are
from~\cite{protoadapter}, which reports 11-dataset means for RN50 only.
$\Delta$ is matched minus published.}
\label{tab:published_vs_matched}
\small
{\setlength{\tabcolsep}{3pt}%
\begin{tabular}{llrrr}
\toprule
Backbone & Method & Published & Matched & $\Delta$ \\
\midrule
\multirow{6}{*}{RN50}
& Zero-shot CLIP     & 58.77 & 58.81 & $+0.04$\\
& Tip-Adapter        & 70.32 & 70.20 & $-0.12$\\
& Proto-Adapter      & 72.69 & 72.48 & $-0.21$\\
& Tip-Adapter-F      & 75.83 & 75.50 & $-0.33$\\
& Proto-Adapter-F    & 75.59 & 75.39 & $-0.20$\\
& GDA-CLIP           & 76.05 & 75.56 & $-0.49$\\
\midrule
\multirow{3}{*}{RN101}
& Zero-shot CLIP     & 59.86 & 59.59 & $-0.27$\\
& Tip-Adapter        & 71.96 & 72.06 & $+0.10$\\
& GDA-CLIP           & 78.06 & 77.41 & $-0.65$\\
\midrule
\multirow{3}{*}{ViT-B/32}
& Zero-shot CLIP     & 61.88 & 62.11 & $+0.23$\\
& Tip-Adapter        & 72.37 & 72.88 & $+0.51$\\
& GDA-CLIP           & 78.06 & 77.69 & $-0.37$\\
\midrule
\multirow{3}{*}{ViT-B/16}
& Zero-shot CLIP     & 65.23 & 65.43 & $+0.20$\\
& Tip-Adapter        & 76.81 & 76.88 & $+0.07$\\
& GDA-CLIP           & 81.85 & 81.40 & $-0.45$\\
\bottomrule
\end{tabular}
}
\end{minipage}
\hfill
\begin{minipage}[t]{0.40\textwidth}
\centering
\captionsetup{hypcap=false}
\captionof{table}{Published GPCache accuracy versus LE-GDA at $K=16$, averaged over 11
datasets.  Published values are from~\cite{gpcache} and include similarity
calibration, which uses external unlabeled images; our GPCache (W+C) baseline
in the main paper's Table~1 does not.  The two columns therefore use different
protocols and evaluation splits, and this table is an external consistency
check rather than a matched comparison.  Our ImageNet accuracies use the
disjoint 40{,}000-image partition of the main paper's Sec.~4 and enter the
LE-GDA means.}
\label{tab:gpcache-published}
\small
\begin{tabular}{lcc}
\toprule
Backbone & GPCache (published) & LE-GDA \\
\midrule
RN50     & 76.75 & 77.47 \\
RN101    & 78.58 & 79.68 \\
ViT-B/32 & 78.49 & 78.69 \\
ViT-B/16 & 82.01 & 82.31 \\
\bottomrule
\end{tabular}
\end{minipage}
\end{strip}

Every deviation in \cref{tab:published_vs_matched} is at most 0.65 points.
The Tip-Adapter and Tip-Adapter-F rows are produced by the authors' released
code.  The Proto-Adapter rows are our implementation; their agreement with
the published RN50 means supports the correctness of our reimplementation.

For RN50 in \cref{tab:gpcache-published}, excluding ImageNet from both means
removes the split mismatch.
Published GPCache then scores 78.00 and LE-GDA scores 78.94, a $+0.94$-point
difference.

\FloatBarrier

\section{Error overlap among selected feature spaces}

We compare the errors made by the classifiers selected by LE-GDA with those of
GDA-CLIP (final-feature).  For the distinct feature spaces in each selected
multiset, we compute
\begin{equation}
O=\operatorname{median}_{i<j}
\frac{\Pr(i\text{ wrong},j\text{ wrong})}
{\Pr(i\text{ wrong})\Pr(j\text{ wrong})}.
\end{equation}
This ratio equals one when the two classifiers' errors are independent.  We
define $H$ as the
fraction of test examples for which at least one selected classifier is correct
while GDA-CLIP (final-feature) is wrong.
The matched search control uses the same number of GDA-CLIP (final-feature) classifiers fitted with
different ridge coefficients.  Every classifier is evaluated at the ensemble's
selected zero-shot fusion coefficient.

\begin{table}[H]
\centering
\caption{Statistics $O$ and $H$ at $K=16$.  $O$ is the median ratio of joint
error probability to the product of the two marginal error probabilities, and
$H$ is the fraction of test examples on which at least one ensemble classifier
is correct while GDA-CLIP (final-feature) is wrong.  Results are averaged over
11 datasets and three random seeds.  ``Selected'' denotes the LE-GDA multiset,
and ``control'' denotes the matched search control, an ensemble containing the
same number of GDA-CLIP (final-feature) classifiers with different ridge
coefficients.}
\label{tab:error_overlap}
\small
\begin{tabular*}{\columnwidth}{@{\extracolsep{\fill}}lrrrrr}
\toprule
Backbone & $O_{\rm sel.}$ & $O_{\rm ctrl.}$ & $H_{\rm sel.}$ & $H_{\rm ctrl.}$ & \shortstack{Distinct\\spaces}\\
\midrule
ViT-B/16 &4.87&6.90&0.0525&0.0221&2.64\\
ViT-B/32 &4.09&6.61&0.0578&0.0242&2.67\\
RN50 &4.08&4.52&0.0764&0.0415&2.55\\
RN101 &4.80&5.56&0.0746&0.0327&2.52\\
\bottomrule
\end{tabular*}
\end{table}

\Cref{tab:error_overlap} shows that, across the 44 dataset--backbone settings,
$H$ is
$0.0352\pm0.0265$ higher for the selected feature spaces than for the matched
search control (41 of 44 positive, $p=1.6\times10^{-9}$).  The selected feature spaces
also have lower $O$ by a median 0.697, but this difference is strongest on the
ViTs.  Crucially, lower $O$ does not
predict the LE-GDA improvement: the pooled Spearman correlation between
$-O$ and the gain is $\rho=0.221$ ($p=0.173$), with backbone-specific
correlations of $0.209,-0.073,0.381,$ and $0.048$ for ViT-B/16, ViT-B/32,
RN50, and RN101.  Thus the selected feature spaces make different errors from
the matched search control, but the analysis does not establish error decorrelation as
the mechanism behind LE-GDA's accuracy gain.

\section{Numerical fidelity to the published GDA-CLIP estimator}

As a numerical fidelity check on DTD with ViT-B/16 at $K=16$, evaluating our
trace-scaled ridge estimator at the coefficient corresponding to the published
GDA-CLIP estimator produced discriminant scores that differed from its
closed-form expression only by the positive factor 0.6818, with identical
argmax predictions on every test image.  This checks the estimator
implementation used by GDA-CLIP (final-feature); it is not presented as a
reproduction of the aggregate accuracies reported for published GDA-CLIP.

\section{Gaussian-mixture ablation protocol}
\label{sec:mda_protocol}

The Gaussian-mixture ablation represents each class by $m$ components in each
feature space.  Components are obtained by clustering the $K$ training
features of each class into $m$ groups after transformation with the fitted
Mahalanobis metric.  The shared covariance is estimated once per feature
space from class-mean residuals and is then held fixed for every value of $m$;
the comparisons therefore do not refit the covariance as the component count
changes.  We evaluate fixed settings $m\in\{1,2,4,8,16\}$ and separately
select $m$ on validation from $\{1,2,4,8\}$, independently for each feature
space.  The analysis covers 10 datasets, four backbones, three independently
sampled $K$-shot training sets, and $K=16$; ImageNet is excluded.  At $m=1$,
the estimator is identical to LE-GDA.

\begin{table}[H]
\centering
\caption{Gaussian-mixture ablation at $K=16$ over 10 datasets, four backbones,
and three independently sampled $K$-shot training sets; ImageNet is excluded.
Mean accuracy is averaged over the 10 datasets and four backbones.  The shared
covariance is estimated once per feature space and held fixed across $m$.
Validation selection searches $m\in\{1,2,4,8\}$; $m=16$ is evaluated only as
a fixed setting.  The $m=1$ estimator is identical to LE-GDA.}
\label{tab:mda_results}
\small
\begin{tabular*}{\columnwidth}{@{\extracolsep{\fill}}lrr}
\toprule
Components & Mean accuracy & $\Delta$ vs. $m=1$\\
\midrule
$m=1$ &80.85&0.000\\
$m=2$ &80.75&$-0.100$\\
$m=4$ &80.47&$-0.378$\\
$m=8$ &79.75&$-1.097$\\
$m=16$ &78.75&$-2.098$\\
Validation-selected $m$ &\textbf{80.87}&+0.020\\
\bottomrule
\end{tabular*}
\end{table}

\Cref{tab:mda_results} shows a monotonic accuracy decrease as the fixed number
of mixture components increases: relative to $m=1$, the loss reaches 2.10
points at $m=16$.  Validation selection recovers only 0.020 points over
$m=1$.  These results support the main-paper conclusion that LE-GDA benefits
from combining feature spaces rather than from adding within-class mixture
components.

\section{Score scaling in the non-standardized control}
\label{sec:supp_scale}

This section gives the full definition of the scale factor $r_\ell$ used by
the non-standardized control in the main paper.

The control replaces the standardized score vector $q_\ell(x)$ with
$\widehat g_\ell(x)=g_\ell(x)/r_\ell$.  The implementation computes
\begin{equation}
r_\ell=\max\Bigl\{\bigl|\operatorname{median}_{u\in\mathcal V}
\textstyle\max_c\, g_{\ell,c}(u)\bigr|,\ \varepsilon_r\Bigr\},
\qquad \varepsilon_r=10^{-12}.
\label{eq:supp_scale}
\end{equation}
The inner quantity is the median over validation images of the largest class
score in space $\ell$.  Two safeguards are applied to it.  The absolute value
handles a negative median.  Without it, dividing by $r_\ell$ would reverse the
class ordering in that space.  The constant $\varepsilon_r$ handles a median
close to zero.  Without it, the scaled scores could be amplified without
bound.  Both are numerical guards.  Neither changes the result when the median
is positive and away from zero.

LE-GDA's standardized fusion does not depend on $r_\ell$.  Per-query
standardization is invariant to positive rescaling of $g_\ell$ apart from its
fixed standard-deviation floor $\sqrt{\delta_v}=10^{-6}$.  Both greedy
selection and the $\alpha$ search read only the standardized scores.

\section{Representation selection under the two fusion rules}
\label{sec:supp_selection}

The representation-selection analysis uses all $11\times4\times3=132$
dataset--backbone--run settings at
$K=16$.  Each backbone therefore contributes 33 dataset--run
combinations.
Selection counts include repeated selections in the multiset.  A selection
rate is such a count divided by 33.  A rate can exceed 100\% because greedy
selection uses replacement.  Under uniform averaging, the selection rates of
the candidate spaces do not order with their standalone accuracy differences
on either ViT backbone.  Under per-query score standardization, they do so on
all four backbones.  The spaces dropped from the uniform rule's selections
have lower dispersion than those retained (median rank difference $-0.333$,
$p=1.6\times10^{-10}$).

\subsection{Fusion rules and controlled selection comparisons}
\label{sec:supp-fusion}

\Cref{tab:supp-fusion-rules} reports seed-level paired counts for LE-GDA, the
matched search control, and every evaluated fusion rule.  Fusion rules are
compared with uniform averaging of globally scaled discriminant scores.
\Cref{tab:supp-dispersion} gives the two controlled comparisons summarized in
the main paper's score-standardization ablation.

\begin{table}[H]
\centering
\caption{Paired comparisons at $K=16$ over 11 datasets, four backbones, and
three random seeds.  The first two rows use GDA-CLIP (final-feature) as the
baseline; the remaining rows use uniform averaging of globally scaled scores.
Zero differences are included in the medians.}
\label{tab:supp-fusion-rules}
\footnotesize
{\setlength{\tabcolsep}{2pt}%
\begin{tabular}{@{}p{0.43\columnwidth}rrrr@{}}
\toprule
Comparison & Median $\Delta$ & Positive & Negative & Tied\\
\midrule
LE-GDA &$+0.829$&109&22&1\\
Matched search control &$+0.000$&52&54&26\\
\midrule
Per-query score standardization &$+0.373$&98&31&3\\
Within-query score ranks &$-0.628$&25&93&14\\
Median across feature spaces &$-0.122$&13&77&42\\
Mean softmax probabilities &$-0.069$&52&67&13\\
\bottomrule
\end{tabular}
}
\end{table}

\Cref{tab:supp-fusion-rules} shows that LE-GDA improves the matched baseline in
109 of 131 non-tied comparisons, whereas the matched search control is nearly
balanced.  Per-query score standardization is the only evaluated fusion
alternative with a positive median difference from uniform averaging.  Rank
averaging, the median across standardized feature spaces, and mean softmax
probabilities all reduce median accuracy.  Thus the conclusions are not
artifacts of averaging the random-seed runs, and scale invariance alone is not
sufficient; the specific standardization rule matters.

\begin{table}[H]
\centering
\caption{Controlled comparisons separating score dispersion from standalone
accuracy at $K=16$.  Here $d$ is the median validation-set dispersion of the
class scores within a query, and Standalone $\Delta$ is the top-1 accuracy
difference relative to matched GDA-CLIP (final-feature)~\cite{gdaclip}.
Counts include repeated selections over 33 dataset--run combinations per
backbone and may exceed 33 because greedy ensemble selection uses replacement.}
\label{tab:supp-dispersion}
\footnotesize
{\setlength{\tabcolsep}{1.5pt}%
\begin{tabular*}{\columnwidth}{@{\extracolsep{\fill}}llccrr@{}}
\toprule
Comparison & Feature & $d$ & Stand.\ $\Delta$ & Unif. & Std.\\
\midrule
\multirow{2}{*}{Dispersion matched}
 & RN50 stage 1 &$4.0\!\times\!10^{-5}$&$-8.09$&10&0\\
 & RN50 stage 3 &$6.3\!\times\!10^{-5}$&$-0.56$&14&44\\
\midrule
\multirow{2}{*}{Accuracy matched}
 & RN101 stage 3 &$1.3\!\times\!10^{-4}$&$+0.48$&24&46\\
 & RN101 final &$3.2\!\times\!10^{-1}$&$0.00$&32&47\\
\bottomrule
\end{tabular*}}
\end{table}

In the dispersion-matched RN50 comparison, stages 1 and 3 have nearly equal
$d$ but differ by 7.53 accuracy points.  Their selection counts under uniform
averaging differ by only four (10 versus 14), whereas per-query score
standardization selects stage~1 zero times and stage~3 44 times.  In the accuracy-matched RN101
comparison, stage~3 is only 0.48 points better than the final embedding
despite a score dispersion more than three orders of magnitude smaller;
uniform averaging selects the two representations 24 and 32 times, whereas
per-query score standardization selects them at nearly equal rates (46 and 47 times).
Together, the two comparisons show that selection under uniform averaging cannot reliably
separate predictive quality from residual score scale.

\subsection{Per-dataset selection under per-query standardization}

\begin{figure*}[p]
\centering
\includegraphics[width=0.93\textwidth]{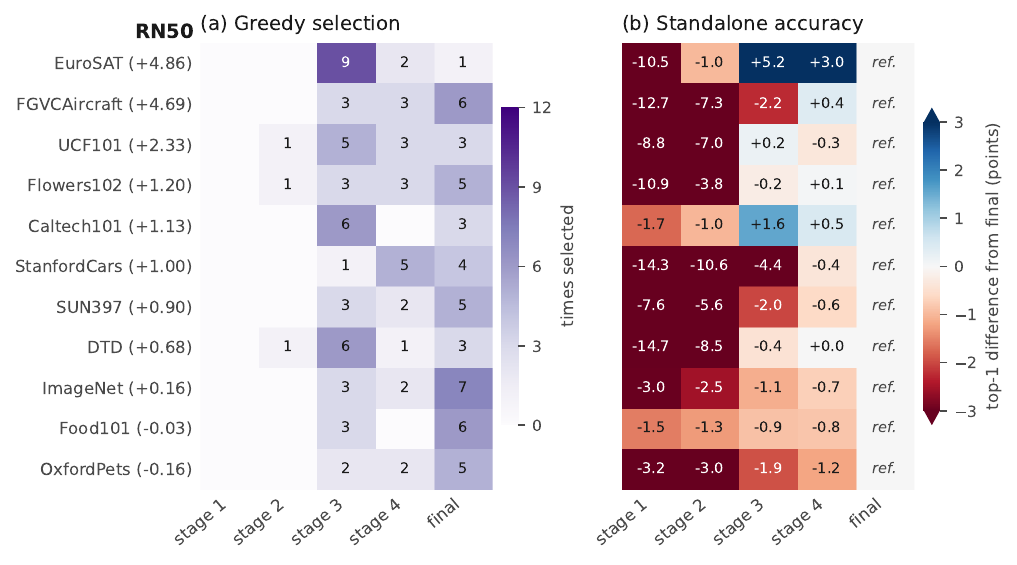}
\caption{Per-query standardized selection counts and standalone top-1
accuracy differences (points) relative to the final embedding for RN50 at
$K=16$. Panel (a) shows greedy selections; panel (b) shows standalone accuracy
differences. Dataset rows are ordered by LE-GDA gain averaged over the four
backbones. Counts are over three runs and include repeated selections; empty
cells denote spaces never selected. Values in parentheses are the gains in
Tab.~2. Colors in (b) are clipped at $\pm3$ points; ref. marks the final
embedding.}
\label{fig:sel_rn50}
\vspace{0.08in}
\includegraphics[width=0.93\textwidth]{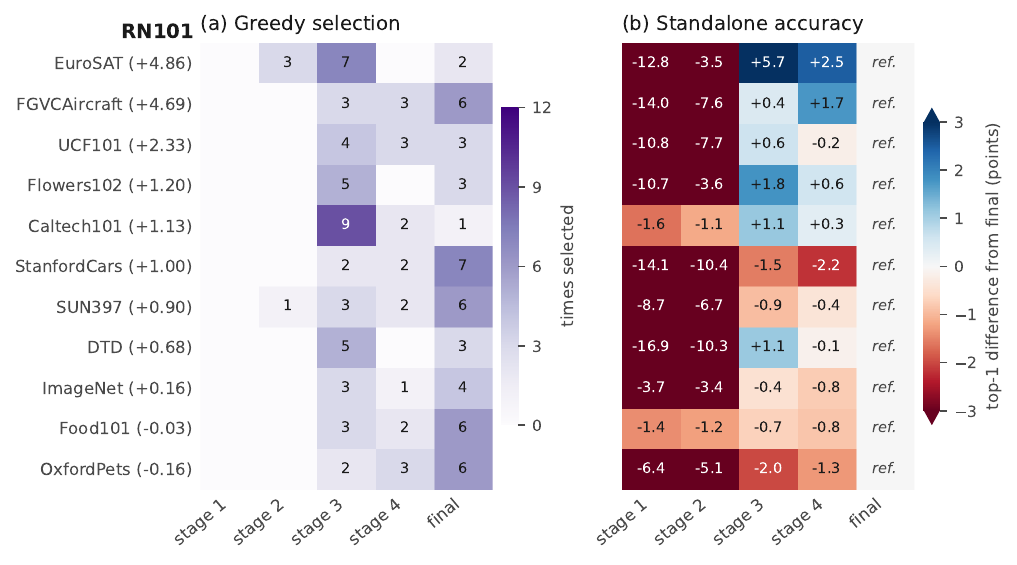}
\caption{Per-query standardized selection counts and standalone top-1
accuracy differences (points) relative to the final embedding for RN101 at
$K=16$. Panel (a) shows greedy selections; panel (b) shows standalone accuracy
differences. Dataset rows are ordered by LE-GDA gain averaged over the four
backbones. Counts are over three runs and include repeated selections; empty
cells denote spaces never selected. Values in parentheses are the gains in
Tab.~2. Colors in (b) are clipped at $\pm3$ points; ref. marks the final
embedding.}
\label{fig:sel_rn101}
\end{figure*}

\begin{figure*}[p]
\centering
\includegraphics[width=0.93\textwidth]{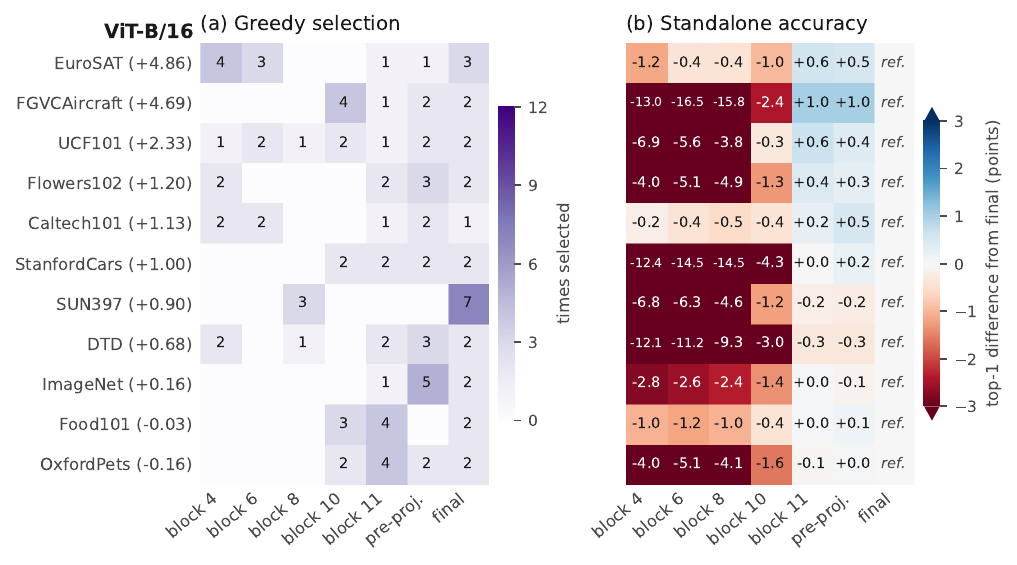}
\caption{Per-query standardized selection counts and standalone top-1
accuracy differences (points) relative to the final embedding for ViT-B/16
at $K=16$. Panel (a) shows greedy selections; panel (b) shows standalone
accuracy differences. Dataset rows are ordered by LE-GDA gain averaged over
the four backbones. Counts are over three runs and include repeated selections;
empty cells denote spaces never selected. Values in parentheses are the gains
in Tab.~2. Colors in (b) are clipped at $\pm3$ points; ref. marks the final
embedding.}
\label{fig:sel_vitb16}
\vspace{0.08in}
\includegraphics[width=0.93\textwidth]{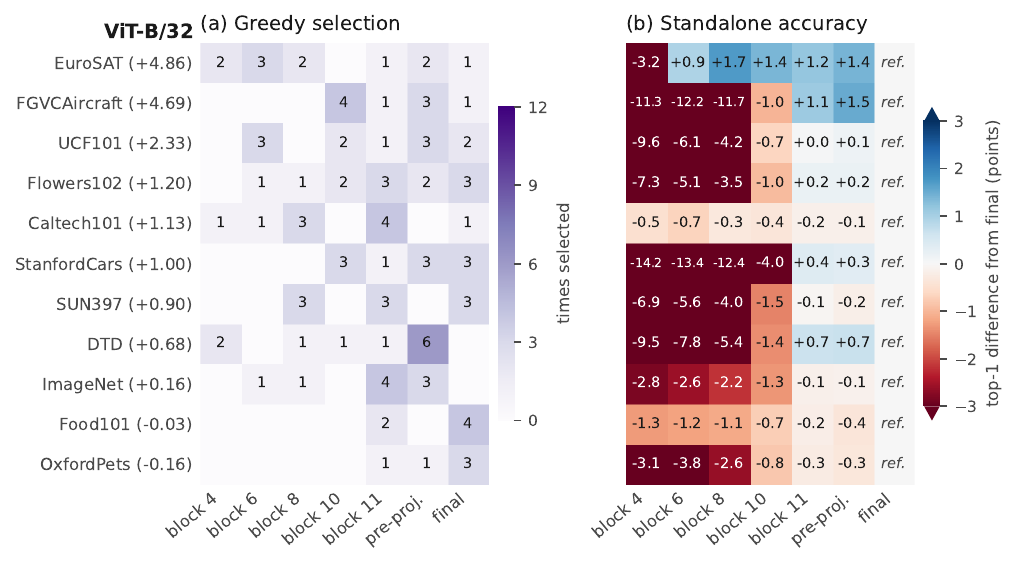}
\caption{Per-query standardized selection counts and standalone top-1
accuracy differences (points) relative to the final embedding for ViT-B/32
at $K=16$. Panel (a) shows greedy selections; panel (b) shows standalone
accuracy differences. Dataset rows are ordered by LE-GDA gain averaged over
the four backbones. Counts are over three runs and include repeated selections;
empty cells denote spaces never selected. Values in parentheses are the gains
in Tab.~2. Colors in (b) are clipped at $\pm3$ points; ref. marks the final
embedding.}
\label{fig:sel_vitb32}
\end{figure*}

Figures~\ref{fig:sel_rn50}--\ref{fig:sel_vitb32} break down the per-query
standardization counts by dataset at $K=16$.  They also report the standalone
top-1 accuracy of each space relative to the final embedding, as in
Tab.~S18.  Rows are ordered by the LE-GDA gain in Tab.~2, averaged over the
four backbones.

On both ResNets, selection follows standalone accuracy.  Stage~1 is never
selected.  On EuroSAT and Caltech101, stages~3 and~4 both exceed the final
embedding, and stage~3 is selected most often.  On ImageNet, Food101, and
OxfordPets, every ResNet intermediate space is less accurate than the final
embedding.  There, the final embedding is selected most often.

Selection on the ViT backbones is more diffuse.  Block~11 and the pre-projection
feature lie within 1.5 points of the final embedding on every dataset.  Greedy
selection therefore spreads its picks across several nearly equivalent
spaces.  On ViT-B/32, for example, the final embedding is never selected on
DTD or ImageNet.

The standalone margins also relate to where LE-GDA improves accuracy.  On the
four datasets with the largest gains, at least one intermediate space exceeds
the final embedding on every backbone.  The clearest case is EuroSAT, where
ResNet stage~3 exceeds the final embedding by more than five points.  On
OxfordPets and Food101, no intermediate space exceeds the final embedding by
more than 0.1 points.  These are the two datasets on which LE-GDA does not
improve GDA-CLIP (final-feature).  Across datasets, the gain correlates with
the best intermediate margin averaged over backbones (Spearman $\rho = 0.87$,
$p < 0.001$).  This association is descriptive and post hoc.

A competitive intermediate space is not required, however.  On SUN397, every
intermediate space is less accurate than the final embedding, yet LE-GDA
improves accuracy by 0.90 points.  On FGVC-Aircraft, the gain of 4.69 points
exceeds every standalone margin, which is at most 1.7 points.  These cases
are consistent with the gain arising from combining representations.

\renewcommand{\dblfloatpagefraction}{0.8}

\clearpage
\renewcommand{\dblfloatpagefraction}{0.99}

\section{Configuration transfer across datasets}

This experiment measures whether LE-GDA's validation-selected configuration
can be transferred to a new dataset at $K\in\{4,16\}$.  For each backbone and
held-out dataset, the representation multiset, ridge coefficients, and fusion
coefficient are derived from the other ten datasets and their three
random-seed runs, then applied unchanged.  The validation-free GDA-CLIP
(final-feature) baseline combines closed-form Ledoit--Wolf shrinkage with a
transferred fusion coefficient, whereas validation-free LE-GDA transfers all
three configuration choices.  Additional configurations restore validation
selection for individual components to quantify their contributions.
\Cref{tab:transfer_configs} defines the six configurations.  Configurations~3
and~1 form the primary validation-free comparison; Configurations~4 and~3 measure the total
contribution of labeled validation access.

\begin{table*}[t]
\centering
\caption{Configurations used in the transfer experiment.  ``Transferred'' denotes a
leave-one-dataset-out value derived from the other ten datasets.}
\label{tab:transfer_configs}
\small
\begin{tabular}{cllll}
\toprule
Configuration & Representation multiset & Ridge coefficient $\lambda$ & Fusion coefficient $\alpha$ & Held-out validation used?\\
\midrule
0 & none (zero-shot CLIP) & none & none & no\\
1 & final representation & closed-form Ledoit--Wolf & transferred & no\\
2 & transferred & validation-selected & validation-selected & yes\\
3 & transferred & transferred & transferred & no\\
3a & transferred & transferred & validation-selected & yes\\
4 & validation-selected & validation-selected & validation-selected & yes\\
\bottomrule
\end{tabular}
\end{table*}

The configurations in \cref{tab:transfer_configs} isolate the three quantities
selected by LE-GDA.  Configuration~3 is fully transferred and uses no held-out
validation labels, while Configurations~2, 3a, and 4 restore validation
selection for the ridge coefficient, fusion coefficient, or representation
multiset.  This makes each paired comparison attributable to a specific source
of validation dependence.

\subsection{Complete configuration results}

\Cref{tab:transfer_k16,tab:transfer_k4} report all six configurations at
16 and four shots.  Accuracy standard deviations are computed across the 11
dataset means for each backbone and therefore describe between-dataset
variation.  Standard deviations of paired contrasts are instead computed
over the 44 dataset--backbone differences.

\begin{table*}[t]
\centering
\caption{Configuration-transfer results at $K=16$.  Entries for Configurations
0, 1, 2, 3, 3a, and 4
are mean accuracy $\pm$ sample standard deviation across the 11 datasets.
The final columns are paired differences between the validation-free configurations and
between fully validation-selected and fully transferred LE-GDA.}
\label{tab:transfer_k16}
\footnotesize
{\setlength{\tabcolsep}{1.5pt}%
\begin{tabular}{lrrrrrrrr}
\toprule
Backbone & Config. 0 & Config. 1 & Config. 2 & Config. 3 & Config. 3a & Config. 4 & $\Delta(3-1)$ & $\Delta(4-3)$\\
\midrule
ViT-B/16 &65.43$\pm$20.48&80.29$\pm$15.16&81.12$\pm$14.45&80.38$\pm$14.43&81.05$\pm$14.44&82.31$\pm$13.55&+0.09$\pm$1.90&+1.93$\pm$1.86\\
ViT-B/32 &62.11$\pm$20.79&76.42$\pm$16.85&77.72$\pm$16.20&76.80$\pm$17.10&77.54$\pm$16.48&78.69$\pm$15.56&+0.39$\pm$1.35&+1.89$\pm$2.34\\
RN50 &58.81$\pm$20.90&73.96$\pm$16.30&76.80$\pm$15.56&75.15$\pm$15.95&76.53$\pm$15.96&77.47$\pm$15.36&+1.19$\pm$3.52&+2.32$\pm$2.16\\
RN101 &59.59$\pm$22.53&75.96$\pm$16.59&78.74$\pm$14.67&76.74$\pm$16.16&78.14$\pm$15.84&79.68$\pm$14.55&+0.78$\pm$3.80&+2.94$\pm$2.81\\
\bottomrule
\end{tabular}}
\end{table*}

\begin{table*}[t]
\centering
\caption{Configuration-transfer results at $K=4$, using the same notation
and dispersion definitions as \cref{tab:transfer_k16}.}
\label{tab:transfer_k4}
\footnotesize
{\setlength{\tabcolsep}{1.5pt}%
\begin{tabular}{lrrrrrrrr}
\toprule
Backbone & Config. 0 & Config. 1 & Config. 2 & Config. 3 & Config. 3a & Config. 4 & $\Delta(3-1)$ & $\Delta(4-3)$\\
\midrule
ViT-B/16 &65.43$\pm$20.48&74.78$\pm$17.03&76.28$\pm$17.66&75.36$\pm$17.36&76.17$\pm$17.95&77.30$\pm$17.14&+0.58$\pm$1.31&+1.95$\pm$2.04\\
ViT-B/32 &62.11$\pm$20.79&71.12$\pm$18.44&72.84$\pm$18.46&71.60$\pm$18.43&72.83$\pm$18.43&73.15$\pm$18.53&+0.48$\pm$1.40&+1.55$\pm$1.80\\
RN50 &58.81$\pm$20.90&66.88$\pm$18.03&69.85$\pm$19.00&66.73$\pm$18.39&69.42$\pm$19.46&70.87$\pm$18.66&$-0.15\pm4.52$&+4.13$\pm$3.20\\
RN101 &59.59$\pm$22.53&69.42$\pm$18.40&72.25$\pm$18.49&69.58$\pm$18.35&71.72$\pm$19.20&73.35$\pm$18.03&+0.16$\pm$2.62&+3.77$\pm$2.81\\
\bottomrule
\end{tabular}}
\end{table*}

\Cref{tab:transfer_k16,tab:transfer_k4} show that the validation-free difference
between Configurations~3 and~1 is modest on every backbone: at $K=16$ it ranges
from 0.09 to 1.19 points, and at $K=4$ it ranges from $-0.15$ to 0.58 points.
In contrast, the fully validation-selected Configuration~4 consistently
outperforms the fully transferred Configuration~3, showing that target-dataset
validation primarily improves configuration selection rather than the
closed-form classifier itself.

\begin{table*}[t]
\centering
\caption{Paired comparisons between transfer configurations.  Differences are computed
after averaging the three runs within each of the 44
dataset--backbone settings.  Sign-test and Wilcoxon $p$-values are two-sided.}
\label{tab:transfer_contrasts}
\small
\begin{tabular}{clrrrrr}
\toprule
$K$ & Comparison & Mean $\pm$ sd & Median & Wins & Sign $p$ & Wilcoxon $p$\\
\midrule
16 &$3-1$ &+0.611$\pm$2.771&+0.593&28/44&0.096&0.126\\
16 &$4-2$ &+0.941$\pm$1.269&+0.561&36/44&$2.5\!\times\!10^{-5}$&$4.6\!\times\!10^{-8}$\\
16 &$2-3\mathrm{a}$ &+0.283$\pm$0.875&+0.087&34/43&$1.7\!\times\!10^{-4}$&$2.1\!\times\!10^{-3}$\\
16 &$3\mathrm{a}-3$ &+1.046$\pm$1.487&+0.338&30/42&$7.9\!\times\!10^{-3}$&$3.1\!\times\!10^{-5}$\\
16 &$4-3$ &+2.270$\pm$2.276&+1.420&41/44&$1.6\!\times\!10^{-9}$&$1.2\!\times\!10^{-10}$\\
\midrule
4 &$3-1$ &+0.267$\pm$2.698&+0.512&28/44&0.096&0.310\\
4 &$4-2$ &+0.862$\pm$1.242&+0.367&36/44&$2.5\!\times\!10^{-5}$&$1.0\!\times\!10^{-7}$\\
4 &$2-3\mathrm{a}$ &+0.271$\pm$0.708&+0.009&22/44&1.00&0.115\\
4 &$3\mathrm{a}-3$ &+1.716$\pm$2.292&+0.747&31/43&$5.4\!\times\!10^{-3}$&$1.9\!\times\!10^{-5}$\\
4 &$4-3$ &+2.850$\pm$2.687&+2.654&40/43&$3.0\!\times\!10^{-9}$&$5.2\!\times\!10^{-8}$\\
\bottomrule
\end{tabular}
\end{table*}

\Cref{tab:transfer_contrasts} confirms this distinction.  The primary
validation-free comparison, Configuration~3 minus Configuration~1, is not
significant at either shot count, whereas Configuration~4 minus
Configuration~3 is positive in 41 of 44 settings at $K=16$ and 40 of 43 at
$K=4$.  At $K=16$, averaging the four backbones within each dataset gives
$\Delta(3-1)=+0.611\pm2.520$ points, with 7/11 positive datasets
(sign-test $p=0.549$; Wilcoxon $p=0.520$).  The same comparison remains
non-significant at $K=4$.  Thus the conclusion does not depend on treating
the four backbones as independent repetitions.

\subsection{Per-dataset results at 16 shots}

\Cref{tab:transfer_datasets} shows that the transfer result is
dataset-dependent and consistent in sign across backbones.  The four negative
datasets have relatively strong zero-shot text classifiers; on these tasks, the
transferred configuration assigns excessive weight to the visual classifier.

All four backbones have the same sign for every dataset: seven datasets are
positive on all four backbones and four are negative on all four.  EuroSAT
accounts for 86\% of the aggregate $\Delta(3-1)$ sum; after removing it, the
mean over the remaining datasets is $+0.09$ points.

\subsection{Validation-curve sensitivity of the fusion coefficient}

Per-query score standardization substantially reduces the between-dataset
dispersion of the selected fusion coefficient.  Nevertheless, proximity to a
validation optimum does not imply that the value transfers without test-set
loss.  Using the transferred $\alpha$ accounts for 46\% of the 2.270-point
difference between Configurations~4 and~3 at $K=16$ and 60\% of the
corresponding 2.850-point difference at $K=4$.  Validation-curve flatness is
measured around each dataset's own selected
multiset, whereas the transferred coefficient is applied together with a
different multiset.  Consequently, local flatness does not bound the loss
from transferring the complete configuration.

\subsection{Checks for validation leakage and model-selection nesting}

The validation-free configurations were evaluated with validation tensors replaced by
an object that raises an exception on every access.  The following checks were
performed before scoring:
\begin{itemize}
  \item The implementation reproduced the reported per-query score-standardization
  results and selected configurations with maximum absolute
  difference zero.
  \item Rescaling every feature-space score vector by a different positive
  constant left Configuration~3 accuracy unchanged, confirming the expected invariance
  of per-query score standardization.
  \item Restricting Configuration~3 to the final representation and using
  Configuration~1's ridge and fusion coefficients reproduced Configuration~1 exactly.
  \item Every transferred multiset contained multiple feature spaces, and no
  selected fusion coefficient was at the boundary of the search grid.
\end{itemize}

\subsection{Ablation with uniform averaging}

As a counterfactual check, transferring the representation multiset does not
produce a significant difference under uniform averaging at $K=16$, whereas
under LE-GDA's per-query standardization it costs $0.941$ points
(\cref{tab:transfer_contrasts}).

\FloatBarrier
\begin{strip}
\centering
\captionsetup{hypcap=false}
\captionof{table}{Per-dataset configuration-transfer differences at $K=16$.  The four
backbone columns report $\Delta(3-1)$.  The final two columns are mean $\pm$
sample standard deviation across backbones.}
\label{tab:transfer_datasets}
\small
{\setlength{\tabcolsep}{3pt}%
\begin{tabular}{lrrrrrr}
\toprule
Dataset & RN101 & RN50 & ViT-B/16 & ViT-B/32 & Mean $\Delta(3-1)$ & Mean $\Delta(4-3)$\\
\midrule
EuroSAT &+9.27&+8.46&+2.18&+3.17&+5.77$\pm$3.61&+2.57$\pm$1.40\\
FGVC-Aircraft &+2.96&+4.11&+3.10&+0.38&+2.64$\pm$1.59&+6.46$\pm$2.04\\
UCF101 &+1.76&+2.80&+0.77&+1.75&+1.77$\pm$0.83&+0.21$\pm$0.35\\
DTD &+2.30&+2.70&+0.71&+0.79&+1.63$\pm$1.03&$-0.04\pm0.51$\\
Flowers102 &+1.57&+1.57&+1.60&+1.07&+1.45$\pm$0.26&+0.85$\pm$0.31\\
Caltech101 &+1.10&+1.61&+0.16&+0.07&+0.73$\pm$0.75&+0.38$\pm$0.26\\
SUN397 &+1.12&+0.47&+0.35&+0.48&+0.60$\pm$0.35&+0.67$\pm$0.24\\
Food101 &$-1.99$&$-1.37$&$-1.32$&$-0.34$&$-1.26\pm0.68$&+3.59$\pm$1.59\\
ImageNet &$-1.40$&$-1.44$&$-1.03$&$-1.28$&$-1.29\pm0.19$&+2.85$\pm$0.96\\
StanfordCars &$-2.56$&$-0.78$&$-2.64$&$-1.69$&$-1.92\pm0.87$&+3.92$\pm$1.08\\
OxfordPets &$-5.55$&$-5.07$&$-2.88$&$-0.15$&$-3.41\pm2.46$&+3.51$\pm$2.69\\
\bottomrule
\end{tabular}}
\end{strip}

\section{Model storage}

After fitting, the discriminant in feature space $\ell$ is stored as the
linear head $f_\ell(x)^\top H_\ell+b_\ell$, where
$H_\ell=\Sigma_\ell^{-1}[\mu_{1,\ell},\ldots,\mu_{C,\ell}]
\in\R^{D_\ell\times C}$ and
$b_\ell=[b_{1,\ell},\ldots,b_{C,\ell}]^\top\in\R^C$, with
$b_{c,\ell}=-\tfrac12\mu_{c,\ell}^\top\Sigma_\ell^{-1}\mu_{c,\ell}$.
Let $\mathcal U^*$ denote the distinct elements of the selected multiset
$\mathcal L^*$.  The inference-time storage is
\begin{equation}
M_{\mathrm{store}}=
\sum_{\ell\in\mathcal U^*}C(D_\ell+1),
\label{eq:supp_storage}
\end{equation}
where $M_{\mathrm{store}}$ counts stored floating-point values.  Each distinct
space stores $CD_\ell$ weights in $H_\ell$ and $C$ entries in $b_\ell$;
repeated selections do not introduce additional heads.

LE-GDA introduces no trainable parameters and requires a single frozen CLIP
forward pass.  Although the selected multiset can contain four elements including repetitions, the
median number of distinct feature spaces is three.  \Cref{tab:storage}
reports the resulting per-backbone storage.

\begin{table}[H]
\centering
\caption{Median number of stored floating-point values per class at $K=16$.
Tip-Adapter uses sparse integer labels in this accounting.}
\label{tab:storage}
\small
\begin{tabular*}{\columnwidth}{@{\extracolsep{\fill}}lrrr}
\toprule
Backbone & \shortstack{GDA-CLIP\\(final-feature)} & LE-GDA & Tip-Adapter\\
\midrule
RN50 &1,025&3,074&16,400\\
RN101 &513&2,562&8,208\\
ViT-B/32 &513&2,051&8,208\\
ViT-B/16 &513&2,051&8,208\\
\bottomrule
\end{tabular*}
\end{table}

\Cref{tab:storage} shows that LE-GDA stores 2,051--3,074 floating-point values
per class, compared with 513--1,025 for GDA-CLIP (final-feature) and
8,208--16,400 for Tip-Adapter.  LE-GDA therefore uses three to five times the
storage of the final-feature discriminant but only 19--31\% of the Tip-Adapter
cache.  The layer ensemble increases storage, but remains substantially smaller
than an example-level cache.

\end{document}